\documentclass[lettersize,journal]{IEEEtran}
\usepackage{amsmath,amsfonts}
\usepackage{algorithmic}
\usepackage{algorithm}
\usepackage{array}
\usepackage[caption=false,font=normalsize,labelfont=sf,textfont=sf]{subfig}
\usepackage{textcomp}
\usepackage{stfloats}
\usepackage{url}
\usepackage{verbatim}
\usepackage{graphicx}
\usepackage{cite}
\usepackage{multirow}
\usepackage{tabularx} 
\usepackage{xcolor}
\usepackage{makecell}
\usepackage{subcaption} 
\usepackage{adjustbox}
\begin{document}

\title{MS2Edge: Towards Energy-Efficient and Crisp Edge Detection with Multi-Scale Residual Learning in SNNs}

\author{Yimeng Fan$^{*}$,
% <-this % stops a space
Changsong Liu$^{*}$\thanks{$*$ These authors contributed equally to this work.},
Mingyang Li,
Yuzhou Dai,
Yanyan Liu,
and 
Wei Zhang$^{\dagger}$\thanks{$\dagger$ Corresponding authors: Wei Zhang (E-mail: tjuzhangwei@tju.edu.cn).}
}



\maketitle

\begin{abstract}
Edge detection with Artificial Neural Networks (ANNs) has achieved remarkable prog\-ress but faces two major challenges. First, it requires pre-training on large-scale extra data and complex designs for prior knowledge, leading to high energy consumption. Second, the predicted edges perform poorly in crispness and heavily rely on post-processing.
Spiking Neural Networks (SNNs), as third generation neural networks, feature quantization and spike-driven computation mechanisms. They inherently provide a strong prior for edge detection in an energy-efficient manner, while its quantization mechanism helps suppress texture artifact interference around true edges, improving prediction crispness.
However, the resulting quantization error inevitably introduces sparse edge discontinuities, compromising further enhancement of crispness.
To address these challenges, we propose MS2Edge, the first SNN-based model for edge detection.
At its core, we build a novel spiking backbone named MS2ResNet that integrates multi-scale residual learning to recover missing boundary lines and generate crisp edges, while combining I-LIF neurons with Membrane-based Deformed Shortcut (MDS) to mitigate quantization errors.
The model is complemented by a Spiking Multi-Scale Upsample Block (SMSUB) for detail reconstruction during upsampling and a Membrane Average Decoding (MAD) method for effective integration of edge maps across multiple time steps.
Experimental results demonstrate that MS2Edge outperforms ANN-based methods and achieves state-of-the-art performance on the BSDS500, NYUDv2, BIPED, PLDU, and PLDM datasets without pre-trained backbones, while maintaining ultralow energy consumption and generating crisp edge maps without post-processing.
\end{abstract}


\section{Introduction}
Edge detection is a longstanding and fundamental task in computer vision, aiming to identify semantically meaningful object boundaries. It is crucial in many high-level applications such as salient object detection \cite{yang_biconnet_2022, su2025rapid}, image segmentation \cite{zhao_edgeenhanced_2025}, and object detection \cite{rani2022object, li_tbnet_2025}. In recent years, edge detection has achieved remarkable progress, with some state-of-the-art (SOTA) methods based on Artificial Neural Networks (ANNs) reaching beyond human-level performance \cite{cetinkaya2024ranked}.

However, these current methods face two major challenges. 
The first is that to acquire prior knowledge, these high-performing edge detection methods typically rely on large-scale extra data pre-training \cite{jing_boosting_2024, zhou2024muge}, as well as complex designs (carefully designed gradient operators \cite{Su_2021_ICCV, liu2025cycle} and multi-model ensemble learning \cite{fu2023practical}). The former leads to significant computational resource waste, while both the inefficient pre-trained backbones and the latter result in substantial energy consumption.
This poses challenges for their real-world applications. 
% Given that edge detection is a foundational task, we believe simpler and more energy-efficient solutions are essential.
The second challenge is that existing edge detection methods achieve high accuracy (accurately distinguishing between edges and non-edges) but lack crispness (precisely locating edge pixels) \cite{ye2023delving, liu2025cycle}.
In other words, their predicted edges often exhibit excessive thickness, encompassing the correct edges but necessitating post-processing to obtain the final results.

\begin{figure*}[t]
\centering
\includegraphics[width=0.8\textwidth]{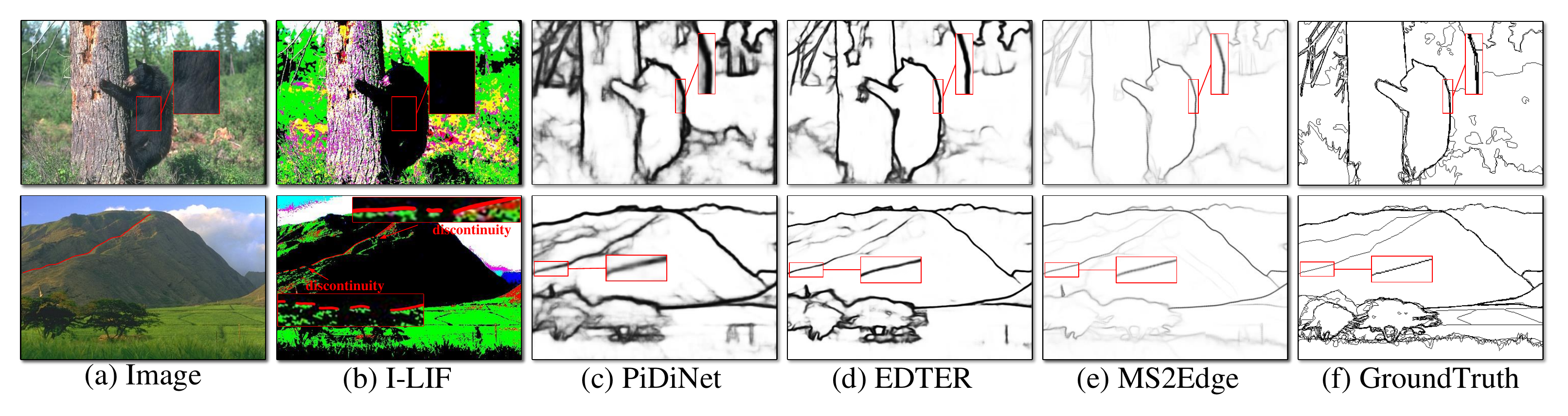}
\caption{\textbf{Visualization of the quantization mechanism in SNNs and the crispness of the proposed MS2Edge.} The images are normalized to the range of 0 to 1 and then directly fed into I-LIF spiking neuron to demonstrate the quantization mechanism of SNNs. For the I-LIF neuron, T is set to 1 and D to 4.}
\label{fig_1}
\end{figure*}

Spiking Neural Networks (SNNs) offer a promising solution to these problems. As the third generation of neural networks, SNNs compute and transmit information using spike signals \cite{tang_ac2as_2023a}. This spike-driven computation mechanism confines operations within the network to simple additions, thereby significantly reducing energy consumption.
Moreover, the quantization mechanism of SNNs converts continuous brightness values into discrete spikes, which enhances the contrast between regions of different brightness magnitudes. This characteristic makes SNNs highly responsive to significant brightness changes at object boundaries while suppressing subtle brightness variations in texture, as shown in Fig. \ref{fig_1}.
{These properties have attracted SNNs to various low-level vision tasks such as image classification \cite{hu2024advancing} and image denoising \cite{castagnetti2023spiden}. 
Unlike these applications that primarily target energy efficiency, SNNs in edge detection not only reduce energy consumption but also address inherent challenges specific to this task.}
As a result, the inherent mechanism of SNNs naturally provides a strong prior for edge detection. {While quantization approaches in ANNs can exhibit similar effects, they require pre-trained models and incur substantial computational costs \cite{MQBench}. In contrast, SNNs can be trained from scratch, thereby fully exploiting the quantization prior.}
This enables SNNs to achieve superior energy-efficient edge detection using a simpler design without pre-trained backbones.

The quantization mechanism in SNNs suppresses false edges caused by textures around true edges, effectively improving the edge localization ability. 
However, this mechanism inevitably introduces quantization errors. In edge detection, these errors excessively suppress boundary lines with subtle brightness changes, leading to sparse edge discontinuities, as shown in Fig. \ref{fig_1}. 
The model can recover these edges based on semantic information, but the absence of fine details like boundary lines leads to high uncertainty, resulting in thick edges. Therefore, this constrains SNNs from achieving further improvements in prediction crispness.
% SNNs exhibit both advantages and limitations in addressing the crispness issue.

To leverage the inherent advantages of SNNs in addressing the issues above while overcoming their limitations, we propose MS2Edge, \textbf{the first attempt to apply SNNs for edge detection}, employing a simple U-shaped architecture.
The key component of our method is the Spiking \textbf{M}ulti-\textbf{S}cale \textbf{M}embrane-\textbf{S}hortcut ResNet (MS2ResNet).
The MS2ResNet employs parallel convolutions with different kernel sizes to capture multi-scale features in the residual path, forming a multi-scale residual learning strategy. 
In this strategy, large-kernel convolutions utilize long-range contextual information to preserve semantic continuity, while small-kernel convolutions precisely capture subtle gradient variations around discontinuous edges, helping to detect cues of missing boundary lines and increase prediction confidence.
By incorporating these multi-scale convolutions into residual learning, which learns perturbations to identity mapping, we can better prevent further information loss during feature extraction.
This effectively addresses the sparse edge discontinuities and enhances the crispness of predicted edges.

Furthermore, to mitigate quantization errors and reduce the possibility of edge discontinuities, we introduce a novel combination of the Membrane-based Deformed Shortcut (MDS) \cite{fan_spikssd_2025} and I-LIF neurons \cite{luo2025integer} within the MS2ResNet. 
The MDS stabilizes the distribution of membrane synaptic inputs across layers. I-LIF neurons provide an effective solution for quantizing extremely large and small membrane synaptic inputs. 
This synergistic design ensures a robust firing pattern across different inputs, thereby reducing information loss from quantization errors.

The decoder also plays a crucial role in crisp edge detection by gradually upsampling to recover fine details from deep semantic features. However, during this process, convolution operations tend to cause similar responses among neighboring pixels, resulting in thick edges in the recovered edge maps \cite{wangDeepCrispBoundaries2017, cao2020learning}.
To address this issue, we introduce the Spiking Multi-Scale Upsample Block (SMSUB), which employs parallel convolutions with different kernel sizes and dilation rates to capture multi-scale features.
This increases the response differences between adjacent pixels, effectively restoring detailed edge information. 
Furthermore, to better integrate predicted edges from SNNs across multiple time steps in the output layer, we propose the Membrane Average Decoding (MAD) method. This process essentially averages the membrane synaptic inputs, hence the name Membrane Average Decoding.

The main contributions of this work can be summarized as follows:

(1) We propose MS2Edge, the first SNN-based edge detection method that leverages the simple U-shaped architecture. It naturally addresses two existing challenges in edge detection by leveraging the quantization and spike-driven computation mechanisms of SNNs.

(2) We propose MS2ResNet, a spiking backbone employing multi-scale residual learning with MDS and I-LIF to address sparse edge discontinuities caused by quantization errors.

(3) We design a Spiking Multi-Scale Upsample Block (SMSUB) to enhance fine detail reconstruction in the decoder, along with a Membrane Average Decoding (MAD) method at the output layer to accurately decode predicted edge maps throughout the entire time steps.

(4) This method achieves SOTA performance with crisp edges and ultralow energy consumption on BSDS500 \cite{arbelaez2010contour}, NYUDv2 \cite{silberman_indoor_2012}, and BIPED \cite{soria2023dense} datasets without pre-trained backbones.
Furthermore, MS2Edge also achieves SOTA performance on both PLDU and PLDM power line detection datasets \cite{zhang2019detecting}, demonstrating its superior generalization capability.

The rest of this paper is organized as follows. Section \ref{related} reports the related work. Section \ref{preliminaries} provides the preliminaries of SNNs. Section \ref{method} introduces our proposed MS2Edge. We present ablation studies and comparisons with SOTA methods in Section \ref{experiment}. Finally, we conclude in Section \ref{conclusion}.

\section{Related work}
\label{related}
\subsection{Spiking Neural Network} 
SNNs are biologically inspired networks that process information through spikes, offering advantages in energy efficiency. Currently, SNNs employ two training methods, including ANNs-to-SNNs conversion and direct training. 
The ANNs-to-SNNs conversion method aims to reduce errors that arise when switching from the activation functions of ANNs to spiking neurons. 
To achieve this goal, researchers have proposed various normalization methods, novel spiking neuron models, and quantized activation functions \cite{tang_ac2as_2023a, hu2023fast}.
Although the ANNs-to-SNNs conversion method has shown promise, it inherently limits SNNs to relying on ANN-derived features, which restricts the full utilization of SNN unique characteristics and their potential to outperform ANNs.
In contrast, direct training with surrogate gradients to optimize SNNs fully leverages these unique characteristics, demonstrating comparable or superior performance in various domains, such as image denoising \cite{castagnetti2023spiden} and object detection \cite{fan2024sfod, luo2025integer}, etc.
Therefore, we adopt the direct training strategy for our model in this research.

Recent advances in SNN architectures have focused  on residual learning based on direct training methods to construct deep networks. Zheng et al. \cite{zheng2021going} builds a 50-layer network with tdBN but merely replaces ReLU with LIF neurons, neglecting spiking characteristics. MS-ResNet \cite{hu2024advancing} mitigates gradient issues and enables deep network training, but its non-spiking down-sampling shortcuts prevent full SNN implementation. EMS-ResNet \cite{su2023deep} addresses the issues in MS-ResNet but fails to improve performance due to limited consideration of SNN characteristics.
MDS-ResNet \cite{fan_spikssd_2025} finds that membrane-based shortcuts increase synaptic input variance in subsequent neurons, weakening feature extraction. To address this, it introduces the MDS to enhance the model in an SNN-friendly manner.
This paper incorporates multi-scale residual learning into Spiking Residual Networks to address the sparse edge discontinuities introduced by the quantization error inherent to SNNs.

\subsection{Edge Detection}
\label{Edge Detection}
Edge detection is a fundamental task in computer vision that has seen significant developments over the years. Early methods primarily relied on calculating image derivative information, which includes Local Binary Pattern (LBP) \cite{Su_2021_ICCV}, Canny \cite{canny2009computational}, etc.
The advent of Artificial Neural Networks (ANNs) has brought about revolutionary progress in this field, giving rise to many ANN-based methods.
These approaches can be divided into two categories based on their prior knowledge incorporation ways. One category leverages pre-trained weights to incorporate prior knowledge, such as HED \cite{Xie_2015_ICCV}, RCF \cite{Liu_2017_CVPR}, BDCN \cite{he_bidirectional_2019}, EDTER \cite{pu_edter_2022}, RankED \cite{cetinkaya2024ranked}, and MuGE \cite{zhou2024muge}.
The second category employs complex designs to eliminate the need for large-scale extra data pre-training, thereby reducing computational costs. These methods, such as PiDiNet \cite{Su_2021_ICCV} and CPD-Net \cite{liu2025cycle}, integrate traditional image gradient information into modern convolutions for high-efficiency edge detection. Pedger \cite{fu2023practical}, on the other hand, achieves this goal through a multi-model ensemble learning approach.
However, the above ANN-based methods, in their pursuit of prior knowledge through large-scale extra data pre-training and complex designs, incur significant energy consumption, limiting their applications in real-world scenarios. In this paper, we demonstrate that SNNs naturally provide strong priors for edge detection, eliminating the dependency on ANN-based methods while significantly improving edge crispness.

\subsection{Spiking Neurons for Edge Detection}
Several studies have explored the application of spiking neurons in edge detection tasks. Wu et al. \cite{wu2007edge} achieve edge detection by integrating spiking neurons with hand-crafted receptive field layers. Meftah et al. \cite{meftah2010segmentation} develop a clustering method using spiking neurons with Hebbian winner-take-all learning for both image segmentation and edge detection. More recently, Fang et al. \cite{fang2024contour} propose a more effective image contour extraction technique by simulating the visual perception pathway, while incorporating color contrast calculation, spiking neuron encoding, and contrast-adaptive dynamic receptive fields.
However, these methods rely on traditional machine learning techniques and merely utilize spiking neurons without constructing complete Spiking Neuron Networks. To address this gap, we propose MS2Edge, the first SNN-based edge detection method.

\section{Preliminaries of SNNs}
\label{preliminaries}

\subsection{Preliminaries of Spiking Neurons}\label{pre_neuron}
The spiking neurons are the core of the quantization and spike-driven computation mechanism in SNNs. To implement these characteristics, researchers have proposed various SNN neuron models, such as Izhikevich \cite{izhikevich2003simple}, Integrate-and-Fire (LIF) \cite{zheng2021going}, and I-LIF \cite{luo2025integer} models.
Currently, the LIF neuron model offers a balance between biological interpretability and computational efficiency, which has led to its widespread adoption in large-scale spiking neural networks \cite{fan2024sfod}. Its neuron model can be described as follows:
\begin{equation}\label{ILIF1}
\mathbf{u}^{t,j} = \mathbf{h}^{t-1,j} + \mathbf{x}^{t,j-1},
\end{equation}
\begin{equation}\label{LIF}
\mathbf{s}^{t,j} = \Theta \left( \mathbf{u}^{t,j} - V_{th} \right),
\end{equation}
\begin{equation}\label{ILIF3}
\mathbf{h}^{t,j} =  \beta_{decay} \left( \mathbf{u}^{t,j} - V_{th} \odot \mathbf{s}^{t,j} \right).
\end{equation}
Here, $t$ denotes the time step, and $j$ represents the layer. The membrane potential $\mathbf{u}^{t,j}$ integrates the temporal information $\mathbf{h}^{t-1,j}$ and spatial information $\mathbf{x}^{t,j-1}$. $V_{th}$ is the membrane potential threshold that determines whether the neuron fires a spike $\mathbf{s}^{t,j}$. $\Theta$ is the Heaviside step function, where it equals $1$ if $x \geq 0$ and $0$ otherwise. The membrane potential decay factor is denoted by $\beta_{decay}$, while $\odot$ indicates element-wise multiplication. 

\begin{figure}[!t]
\centering
    \includegraphics[width=1.\columnwidth]{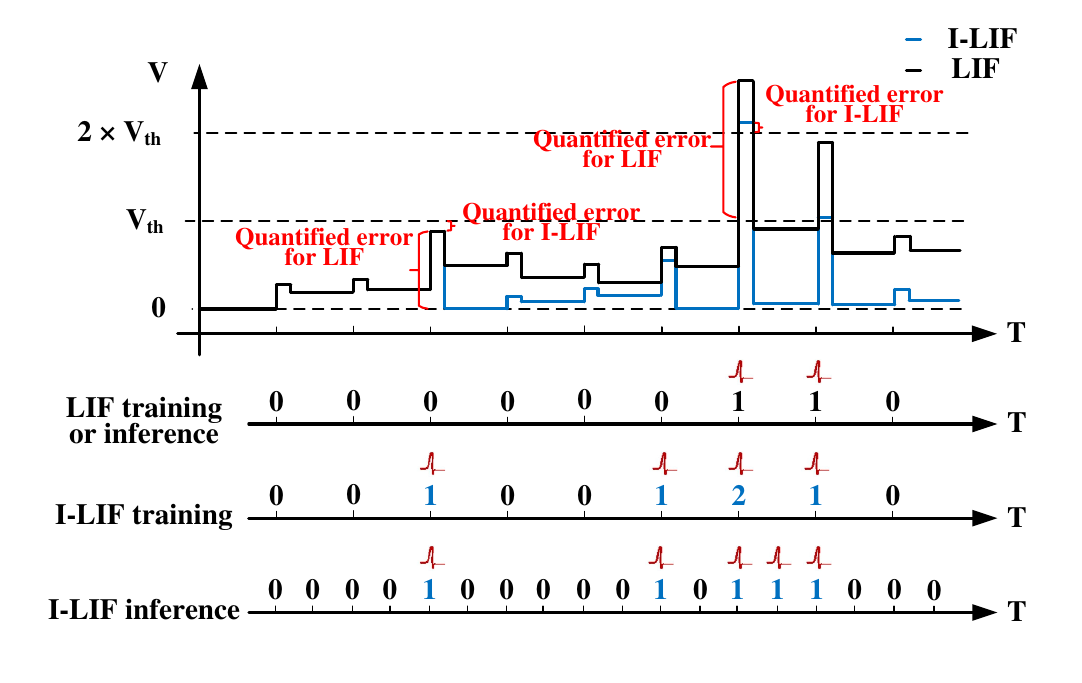}
\caption{\textbf{The quantization error for I-LIF and LIF neurons when processing identical inputs.}}
\label{fig_2}
\end{figure}

However, the LIF neuron introduces significant quantization errors due to its limitation of firing only 0 or 1 spikes. 
To address this issue, Luo et al. \cite{luo2025integer} propose the I-LIF neuron, which modifies Eq. \ref{LIF} into
\begin{equation}\label{ILIF2}
\mathbf{s}^{t,j} = \mathit{Clip} \left( \mathit{round}(\mathbf{u}^{t,j}), 0, D \right),
\end{equation}
to enable the firing of integer-valued spikes.
Here, $\mathit{Clip} \left( x, min, max \right)$ refers to the operation of clipping $x$ within the range $[min, max]$, $\mathit{round}(x)$ denotes the rounding function, and $D$ is a hyperparameter defining the maximum integer value for spike firing.
The I-LIF neuron is trained in this form and ensures the spike-driven computational mechanism by converting the integer values into binary spikes during inference. 
As shown in Fig. \ref{fig_2}, I-LIF neurons effectively reduce quantization error for both small and large input magnitudes.
Given its numerous advantages, our proposed MS2Edge employs I-LIF neurons.

\begin{figure*}[t]
\centering
\includegraphics[width=1.\textwidth]{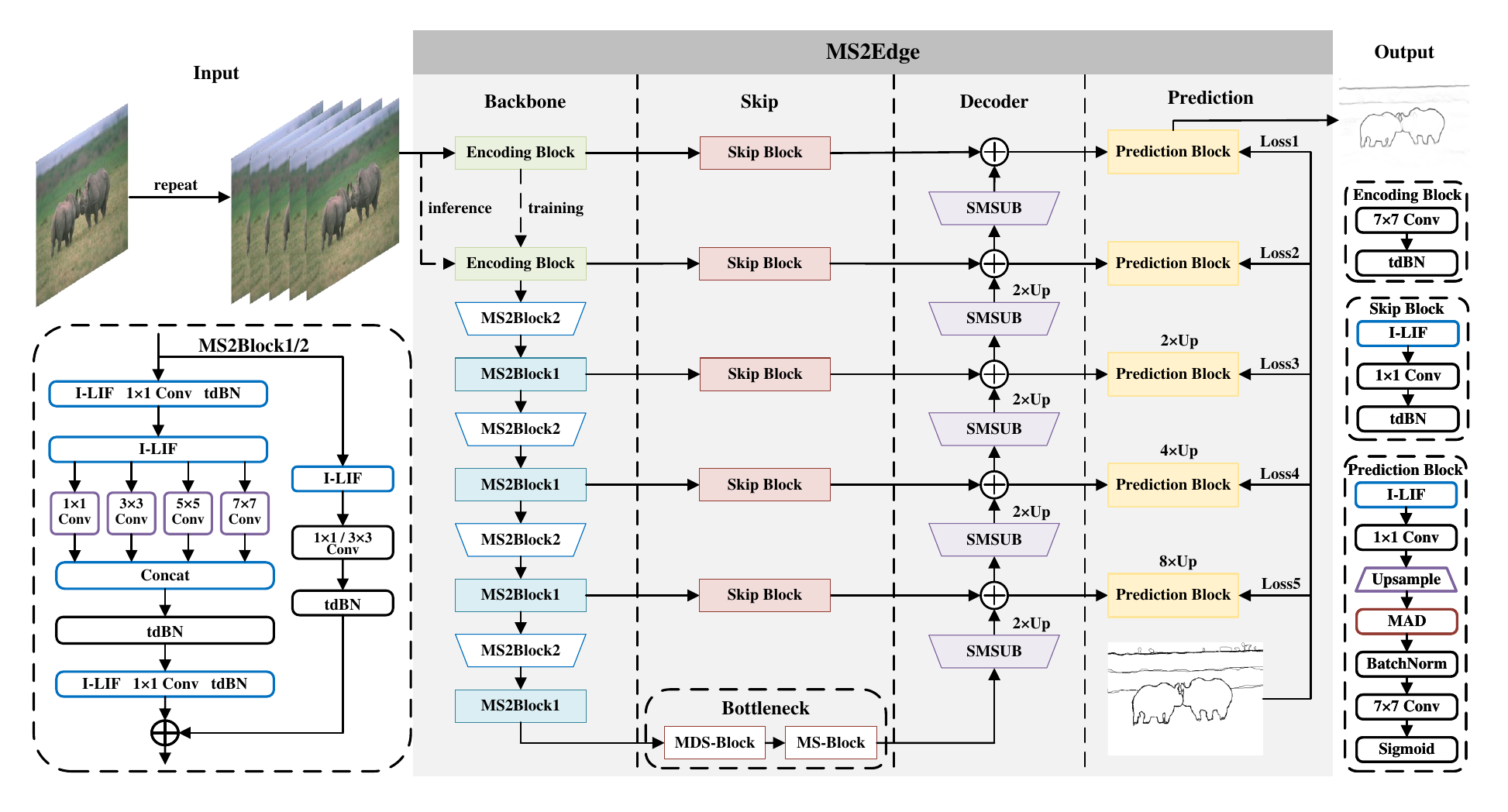}
\caption{\textbf{The architecture of proposed MS2Edge.} It consists of the backbone, bottleneck module, skip module, decoder, and prediction module. 
The backbone uses MS2ResNet26 configuration, detailed in Appendix B.
The MS2ResNet consists of MS2Block1 and MS2Block2, with the latter using $3\times3$ convolution in the MDS for downsampling to better incorporate spatial information. 
In the decoder, we use SMSUB for upsampling and fuse it with backbone features that are processed through the Skip Block. Finally, the Prediction Block is used for prediction. 
}
\label{fig_3}
\end{figure*}

\subsection{Preliminaries of Training Strategy}\label{pre_train}
During SNN training, the non-differentiability occurs in the spike firing process shown in Eqs. \ref{LIF} and \ref{ILIF2}. To address this, surrogate gradients are employed to enable computation within modern deep-learning frameworks. Following \cite{hu2024advancing}, the surrogate gradient for Eq. \ref{LIF} is defined as:
\begin{equation}
    \frac{\partial \mathbf{s}^{t,j}}{\partial \mathbf{u}^{t,j}} = \frac{1}{a}\text{sign}\left(\left|\mathbf{u}^{t,j} - V_{\text{th}}\right| \leq \frac{a}{2}\right),
\end{equation}
where $a$ is a coefficient ensuring the function integrates to 1. Following \cite{luo2025integer}, the surrogate gradient for Eq. \ref{ILIF2} is defined as:
\begin{equation}
    \frac{\partial \mathbf{s}^{t,j}}{\partial \mathbf{u}^{t,j}} = \begin{cases} 
        1, & \text{if } 0 \leq \mathbf{u}^{t,j} \leq D \\
        0, & \text{otherwise.}
    \end{cases}
\end{equation}

\section{Methodology}
\label{method}
\subsection{Overview}
The architecture of the proposed MS2Edge is shown in Fig. \ref{fig_3}.
Firstly, we adopt the direct coding method \cite{fan_spikssd_2025} to convert static RGB images into spike trains. Unlike rate coding, which relies on fixed probability sampling, direct coding employs a learnable encoding layer, enhancing the model learning capacity. To further optimize this process, we deepen the encoding layers to two.
Such improvements do not compromise the spike-driven computation mechanism of the model. During inference, the two encoding layers in the backbone can be reparameterized into a single convolution layer \cite{zheng2021going}, while retaining the original first encoding layer to avoid disrupting the skip module. Fig. \ref{fig_3} illustrates the information flow difference between training and inference.

The spike trains are input into MS2ResNet for multi-scale feature extraction, with the deepest feature maps forwarded to the bottleneck module. This module enhances global semantic feature representation while performing two other critical functions: the MDS-Block stabilizes the membrane synaptic input distribution, and the MS-Block ensures stable gradient propagation in the deepest layer.
The output of the bottleneck is fed to the decoder, where the Spiking Multi-Scale Upsample Block (SMSUB) performs feature map upsampling. It then fuses these upsampled maps with the corresponding scale feature maps from the backbone using membrane-based addition. This approach effectively integrates deep semantic information with shallow spatial details, enhancing the overall feature representation.
Notably, the multi-scale features extracted from the backbone are processed by Skip Blocks before fusion. These blocks not only refine the backbone features but also adjust their output distributions by tdBN.

Finally, MS2Edge generates edge maps at multiple scales through Prediction Blocks for deep supervision, but only the output from the first block is used as the final output. 
Therefore, the Prediction Block, which is not required during inference, can be pruned to further reduce parameters. 
The Membrane Average Decoding (MAD) method is employed to decode information across multiple time steps.

\subsection{MS2ResNet}\label{MS2ResNet}
The multi-scale feature is crucial for edge detection \cite{Xie_2015_ICCV, fu2023practical} because shallow features provide detailed information but lack semantic information, making them ineffective in suppressing non-edge textures. In contrast, deep features exhibit strong semantic information but are deficient in detail. 
Therefore, multi-scale feature extraction and integration can effectively capture complementary information, leading to more precise edge representation.
In edge detection with SNNs, while the quantization mechanism can suppress texture interference, its quantization errors disrupt boundary lines, leading to sparse discontinuities in edges.
When predicting these local edges, these discontinuities may limit the edge crispness, thereby preventing full exploitation of the inherent advantages of SNNs.
Therefore, leveraging multi-scale information is even more critical when using SNNs for edge detection.

Based on the above analysis, we propose a multi-scale residual learning strategy.
Specifically, we first apply a $1\times1$ convolution to exchange information between channels. 
Then, we group the feature maps and process them with convolutions of different kernel sizes. 
In this way, small kernel convolutions like $1\times1$ and $3\times3$ capture local gradient cues related to missing boundary lines. Since the edge discontinuities only occur at sparse boundary locations while the overall semantic structure remains continuous (as shown by the mountain ridges in Fig. \ref{fig_1}), large kernel convolutions like $5\times5$ and $7\times7$ are used to capture long-range semantic context information.
Finally, a $1\times1$ convolution is used to aggregate the information.
As shown in Fig. \ref{fig_3}, by integrating multi-scale information into the residual path, we can continuously supplement missing edge information in the identity mapping while avoiding further disruption of boundary lines from quantization errors.

For quantization errors, it is important to note that they cannot be completely eliminated, as they are inherent to the quantization process. The suppression of textures is essentially a beneficial side effect of this quantization error. Therefore, we introduce the combination of Membrane-based Deformed Shortcut (MDS) and I-LIF in MS2ResNet to mitigate these errors, achieving a balance between edge preservation and texture suppression.
By enabling the shortcut path to adjust the output distribution, MDS effectively provides more stable membrane synaptic input distributions for the subsequent spiking neurons, reducing the probability of excessively large or small inputs.
Nevertheless, the occurrence of extreme inputs, though less frequent, remains possible.
To address this, I-LIF offers a solution as shown in Fig. \ref{fig_2}. For excessively small inputs, the round function in I-LIF provides an effective quantization method. For excessively large inputs, even when reaching twice the threshold, I-LIF neurons can trigger integer values during training and convert these integers into spike trains during inference, providing an effective solution for quantizing such inputs. 
% By combining these two approaches, the model effectively mitigates the quantization error issue and reduces the probability of discontinuous edges.

By combining the multi-scale residual learning with MDS and I-LIF mentioned above, we propose a novel Spiking Residual Network named MS2ResNet, whose structure is shown in the backbone of Fig. \ref{fig_3}. Specifically, it mainly consists of two blocks: MS2Block1 and MS2Block2. MS2Block1 is used for non-downsampling feature extraction, while MS2Block2 is designed for downsampling. 
In MS2Block2, to prevent information loss from 1$\times$1 convolution with stride 2 and better integrate spatial information, we replace the convolution in MDS with a 3×3 convolution with stride 2, while using multi-scale convolutions with stride 2 in the residual path for downsampling.
% As the model goes deeper, we also partially utilize the MS-Block \cite{su2023deep} to facilitate gradient propagation, while preserving the functionality of MS2Block.

\subsection{Analysis of Gradient Vanishing/Explosion for MS2ResNet}
\label{analysis}
To demonstrate that the MS2ResNet can be trained sufficiently deep without gradient vanishing or explosion problems, we employ Block Dynamical Isometry \cite{chen2020comprehensive} to verify the gradient norm equality.

We introduce some preliminaries from \cite{chen2020comprehensive}, which view the network as a series of blocks as shown in Eq. \ref{serial}. Here, \( \mathbf{f_{j, \theta}} \) denotes the j-th block with parameter vector \( \theta \). For simplicity of analysis, they denote the Jacobian matrix of the j-th block as \(\mathbf{J_j} = \frac{\partial \mathbf{f_j}}{\partial \mathbf{f_{j-1}}}\). Let \( \phi (\mathbf{J}) \) denote the expectation of \( \text{tr}(\mathbf{J}) \), additionally, \(\varphi(\mathbf{J})\) is defined as \(\phi(\mathbf{J}^2)-\phi^2(\mathbf{J})\).
\begin{equation}\label{serial}
\mathbf{f(x)}=\mathbf{f_{l, \theta}} \circ \mathbf{f_{l-1, \theta}} \circ \cdots \circ \mathbf{f_{1, \theta}(x)}
\end{equation}
\noindent\textbf{Definition 1.} 
\textbf{(Definition 3.1. in \cite{chen2020comprehensive})}
\textit{Consider a neural network that can be represented as a series of blocks as Eq. (\ref{serial}) and the j-th block’s jacobian matrix is denoted as \(\mathbf{J_j}\). If \(\forall j\), \(\phi(\mathbf{J_j}\mathbf{J_j}^{T})\approx1\) and \(\varphi(\mathbf{J_j}\mathbf{J_j}^{T})\approx0\), the network achieves Block Dynamical Isometry and can avoid gradient vanishing or explosion.} 

\noindent\textbf{Definition 2.}
\textbf{(Definition 5.1. \cite{chen2020comprehensive})} \textit{Let \(\mathbf{f(x)}\) be a transform whose Jacobian matrix is \(\mathbf{J}\). \(\mathbf{f}\) is called general linear transform when it satisfies: } 
\begin{equation}\label{linear}
E\left[\frac{||\mathbf{f(x)}||_2^2}{len(\mathbf{f(x)})}\right] = \phi\left(\mathbf{J}\mathbf{J}^T\right)E\left[\frac{||\mathbf{x}||_2^2}{len(\mathbf{x})}\right].
\end{equation}
\noindent\textbf{Theorem 1.}
\textbf{(Theorem 4.1. in \cite{chen2020comprehensive})}
\textit{Given \(\mathbf{J}:=\prod_{j=L}^{1}\mathbf{J_j}\), where \(\{\mathbf{J_j} \in \mathbb{R}^{m_j\times m_{j-1}}\}\) is a series of independent random matrices. If \((\prod_{j=L}^1\mathbf{J_j})(\prod_{j=L}^1\mathbf{J_j})^T\) is at least the \(1^{st}\) moment unitarily invariant, we have}
\begin{equation}
\phi\left((\prod_{j=L}^1\mathbf{J_j})(\prod_{j=L}^1\mathbf{J_j})^T\right)=\prod_{j=L}^1\phi(\mathbf{J_j}\mathbf{J_j}^T).
\end{equation}

It is worth noting that \(\varphi(\mathbf{J_j}\mathbf{J_j}^{T})\approx0\) ensures that the block can steadily achieve \(\phi(\mathbf{J_j}\mathbf{J_j}^{T})\approx1\). In many cases, \(\phi(\mathbf{J_j}\mathbf{J_j}^{T})\approx1\) is sufficient for analysis.
Based on Definitions 1 and 2, MS2ResNet can be viewed as a series of blocks and verified each block satisfies Block Dynamical Isometry. Then, according to Theorem 1, the entire network can be analyzed.

\begin{figure}[!t]
\centering
    \includegraphics[width=.9\columnwidth]{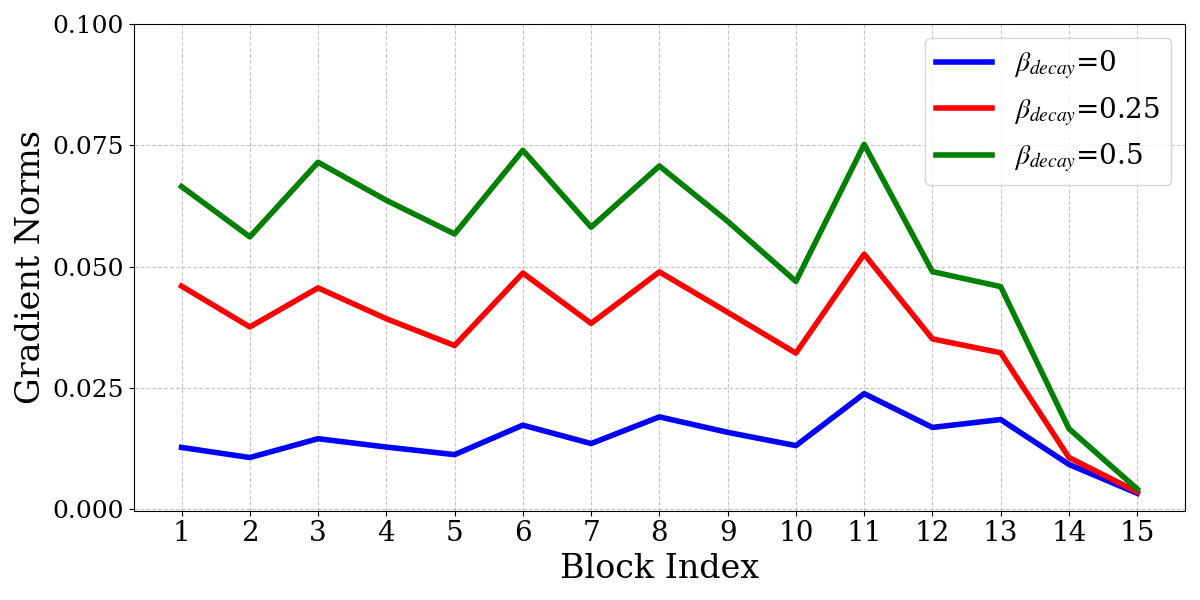}
\caption{\textbf{The visualization of gradient norm in MS2ResNet.}}
\label{fig_4}
\end{figure}

For SNNs, the main difference from ANNs in gradient analysis is that the gradient information propagates not only along the spatial dimension but also along the temporal dimension in spiking neurons. Specifically, let \(\mathbf{L}\) denote the loss function. Based on Eqs. \ref{ILIF1}, \ref{ILIF2}, and \ref{ILIF3} and the chain rule, the gradient of I-LIF neurons can be derived as follows:
\begin{align}\label{neuron1}
\frac{\partial \mathbf{L}}{\partial \mathbf{s}_i^{t,j}} 
% &= \sum_{j=1}^{n_{t,l+1}} \frac{\partial \mathbf{L}}{\partial \mathbf{U}_j^{t,l+1}} \frac{\partial \mathbf{U}_j^{t,l+1}}{\partial \mathbf{S}_i^{t,l}} +
% \frac{\partial \mathbf{L}}{\partial \mathbf{H}_i^{t,l}} \frac{\partial \mathbf{H}_i^{t,l}}{\partial \mathbf{S}_i^{t,l}} \notag \\
&= \sum_{j=1}^{n_{t,j+1}} \frac{\partial \mathbf{L}}{\partial \mathbf{u}_j^{t,j+1}} \frac{\partial \mathbf{u}_j^{t,j+1}}{\partial \mathbf{s}_i^{t,j}} - \beta_{decay} V_{th} \cdot \frac{\partial \mathbf{L}}{\partial \mathbf{h}_i^{t,j}},
\end{align}
\begin{align}\label{neuron2}
\frac{\partial \mathbf{L}}{\partial \mathbf{u}_i^{t,j}} 
% &= \frac{\partial \mathbf{L}}{\partial \mathbf{S}_i^{t,l}} \frac{\partial \mathbf{S}_i^{t,l}}{\partial \mathbf{U}_i^{t,l}} + \frac{\partial \mathbf{L}}{\partial \mathbf{H}_i^{t,l}} \frac{\partial \mathbf{H}_i^{t,l}}{\partial \mathbf{U}_i^{t,l}} \notag \\
&= \frac{\partial \mathbf{L}}{\partial \mathbf{s}_i^{t,j}} \frac{\partial \mathbf{s}_i^{t,j}}{\partial \mathbf{u}_i^{t,j}} + \beta_{decay} \cdot \frac{\partial \mathbf{L}}{\partial \mathbf{h}_i^{t,j}}.
\end{align}

From the above equation, gradient information in the temporal domain is propagated through \(\beta_{decay}\). Since \(\beta_{decay}\) typically takes small values (e.g., 0.25, 0.5), its effect on gradient propagation is negligible. Therefore, for simplicity of analysis, let \(\beta_{decay} = 0\). This leads to the following propositions.

\noindent\textbf{Proposition 1.}
\textit{For MS2Block1 and MS2Block2, if \(\beta_{decay} = 0\) and the block output at each time step $t$ follows \(\mathbf{x^t} \sim \mathcal{N}(0, 1)\), then each block satisfies: \(\phi(\mathbf{J_{j}^{t}(J_{j}^{t})^{T}}) = \frac{1}{\alpha_{2}^{t,j-1}}\).}

\vspace{1pt}

\noindent\textbf{Proposition 2.}
\textit{For MS2ResNet, if all conditions stated in Proposition 1 hold and the $2^{nd}$ moment of the encoding layer output can be controlled at each time step $t$, then \(\phi(\mathbf{J^t(J^t)^T}) \approx 1\). Consequently, the training of MS2ResNet can avoid gradient vanishing or exploding problems.}

\noindent\textbf{Proof} The details can be found in the Appendix A.

To verify that \(\beta_{decay}\) has minimal impact on gradient propagation when \(\beta_{decay} \neq 0\), we visualize the average gradient norm of each block from 256 test images on the CIFAR10 dataset, as shown in Fig. \ref{fig_4}. The network architecture follows ResNet32 from \cite{he_deep_2016}. 
Although the gradient norm gradually increases as \(\beta_{decay}\) increases, this growth does not affect training. Therefore, we can conclude that MS2ResNet effectively prevents gradient vanishing or exploding issues.

\subsection{Spiking Multi-Scale Upsample Block}
The decoder progressively upsamples feature maps to restore spatial details.
However, similar responses of convolution inevitably degrade the recovered details.
In SNNs, we should replace the bilinear interpolation with nearest-neighbor interpolation to maintain the spike-driven computation mechanism, but this creates jagged artifacts along curved edges \cite{parsania2014review}.
Essentially, these jagged artifacts arise from introducing pixels with identical values, lacking smooth edge transitions, which amplifies the similar response problem.

\begin{figure}[!t]
\centering
    \includegraphics[width=1.\columnwidth]{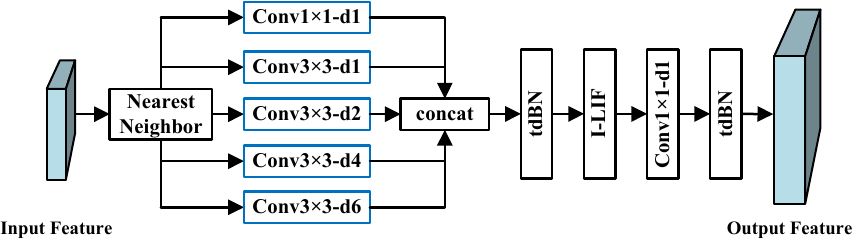}
\caption{\textbf{The architecture of SMSUB}, where d represents the dilation rate.}
\label{fig_5}
\end{figure}

To address this issue, we propose a Spiking Multi-Scale Upsample Block (SMSUB), as illustrated in Fig. \ref{fig_5}. We employ multi-scale convolutions with different kernel sizes and dilation rates after interpolation to incorporate information from edge neighborhoods, which helps generate distinct responses for identical pixels and suppresses edge distortions. Then, SMSUB integrates these features through a 1$\times$1 convolution. 
Finally, this mechanism overcomes the drawbacks of nearest-neighbor interpolation and enhances the model's ability to restore detailed edge information, thus improving edge crispness.

\subsection{Membrane Average Decoding}
\label{MAD}
Currently, the most commonly used spiking decoding methods are Spiking Rate Decoding (SRD) \cite{fan2024sfod} and Last Membrane Potential Decoding (LMPD) \cite{su2023deep}.
Specifically, SRD calculates the spiking rate as the decoding output. While this method aligns with SNN characteristics, its restriction to discrete output values compromises prediction accuracy.
LMPD suppresses spike firing in the final layer and uses the accumulated membrane potential as the final output. Although this method enhances the expressive capacity of the model and improves prediction accuracy, the continuous accumulation of membrane potential for output can amplify subtle edge artifacts, which is harmful to generating crisp edges.

To address these issues, we propose the Membrane Average Decoding (MAD) method. This method modifies the neurons in the final layer as follows:
\begin{equation}
\begin{cases}
\begin{aligned}
\mathbf{u_t} &= \mathbf{u_{t-1}} + \frac{1}{T}\mathbf{x_t}, \\
\mathbf{o} &= \mathbf{u_T},
\end{aligned}
\end{cases}
\end{equation}
where \(\mathbf{u_t}\) and \(\mathbf{x_t}\) represent the membrane potential of the neuron and the input at time step t, respectively, \(\mathbf{o}\) denotes the neuron output, and T indicates the total time steps.

Essentially, this method averages the membrane synaptic input at each time step, which not only considers information from all time steps without continuous membrane potential accumulation but also overcomes the issue of inadequate expressive capacity in SRD.
This enables it to achieve both crispness and accuracy when decoding edge maps across multiple time steps.
Moreover, this approach offers computational simplicity during inference, as it only requires dividing the parameters of the previous layer by T. This allows for addition-only during decoding, thereby eliminating division operations.

\section{Experiments}
\label{experiment}
\subsection{Experimental Setup}
\subsubsection{Datasets and Competitors}
We evaluate MS2Edge on five diverse datasets: BSDS500 \cite{arbelaez2010contour}, NYUDv2 \cite{silberman_indoor_2012}, BIPED \cite{soria2023dense}, PLDU, and PLDM \cite{zhang2019detecting}. BSDS500 contains 200 training images, 100 validation images, and 200 test images, with each image annotated by approximately 5 to 7 annotators. 
During training, we combine BSDS500 with the Pascal VOC Context dataset (containing 10,103 images) to further improve performance.
NYUDv2 consists of 1449 RGB-HHA image pairs, divided into 381 training, 414 validation, and 654 test samples. BIPED is a high-quality outdoor scene dataset comprising 250 high-resolution images, split into 200 training and 50 test images. PLDU contains 453 training images and 120 test images, serving as the primary dataset for power line detection evaluation. Following \cite{zhang2019detecting}, we only use the PLDM test set (50 images) for cross-dataset evaluation due to its relatively small size.

We compare our method with recent SOTA edge detectors, including traditional methods such as Canny \cite{canny2009computational}, gPb-UCM \cite{arbelaez2010contour}, SE \cite{dollar_fast_2014}, and MES \cite{sironi2015projection}, and ANN-based approaches such as HED \cite{Xie_2015_ICCV}, RCF \cite{Liu_2017_CVPR}, CED \cite{wangDeepCrispBoundaries2017}, LPCB \cite{Deng_2018_ECCV}, BDCN \cite{he_bidirectional_2019}, DexiNed \cite{soria2023dense}, PiDiNet \cite{Su_2021_ICCV}, FCL-Net \cite{xuan2022fcl}, EDTER \cite{pu_edter_2022}, PEdger \cite{fu2023practical}, MuGE \cite{zhou2024muge}, RankED \cite{cetinkaya2024ranked}, and CPD-Net \cite{liu2025cycle}. 

\subsubsection{Implementation Details}
% We implement our network using the SpikingJelly \cite{fangSpikingJellyOpensourceMachine2023} deep-learning framework. 
The batch size is 8 for the BSDS500, BIPED, and PLDU datasets, and 4 for NYUDv2. Due to the high resolution of BIPED images, we randomly crop them to $320\times320$ during training. A warm-up strategy is employed for the first 4 epochs, gradually increasing the learning rate to $1e-3$, followed by decay using cosine learning rate scheduling. We use the Adam optimizer and train the model for 30 epochs. 
For data augmentation, we follow \cite{Deng_2018_ECCV} by applying three-way flipping (horizontal, vertical, and both) and 24 different rotation angles across all datasets.
{All experiments are conducted on a server equipped with two NVIDIA Tesla A40 GPUs (48GB each), dual Intel Xeon Silver 4310 processors (2.10GHz, 24 cores), and 512GB of RAM.}
% All experiments are conducted on two Tesla A40 GPUs.

% During training, we follow the common practice \cite{liu2017richer, su2021pixel, zhang2019detecting} to combine the training and validation sets for BSDS500 and NYUDv2, while maintaining the default configuration for BIPED and PLDU.

% Following previous works \cite{deng2018learning}, we augment the training data with the PASCAL VOC Context dataset \cite{mottaghi_cvpr14}, which includes 10,103 images, to further enhance model performance.

\subsubsection{Evaluation Metrics}
To evaluate both the performance and edge map crispness of MS2Edge, we employ widely adopted metrics \cite{ye2023delving} including Standard Evaluation Protocol (S-Eval), Crispness-Emphasized Evaluation Protocol (C-Eval), and Average Crispness (AC). S-Eval calculates ODS (Optimal Dataset Scale) F-score, OIS (Optimal Image Scale) F-score, and AP (Average Precision) after applying NMS and morphological thinning. C-Eval computes F-scores without post-processing, better evaluating the model raw output crispness.
The AC metric evaluates edges through the ratio of pixel sums after and before NMS processing, where higher scores reflect crisper edges.
The localization tolerance is set to 0.0075 for BSDS500, BIPED, PLDU, and PLDM, and 0.011 for NYUDv2 during evaluation.

We also report the energy consumption to quantify the energy efficiency of the network, a metric commonly used in SNNs. SNN energy consumption stems from performing Accumulation Calculations (ACs) only when neurons fire. In addition, some SNN models \cite{hu2024advancing} involve Multiplication and Addition Calculations (MACs) due to suboptimal design, so we include these in our calculations.
For ANNs, we focus on MACs, as they dominate. 
{Following \cite{luo2025integer, horowitz20141}, we use $E_\text{MACs}=4.6\text{pJ}$, $E_\text{ACs}=0.9\text{pJ}$ for FLOAT32 and $E_\text{MACs}=0.23\text{pJ}$, $E_\text{ACs}=0.03\text{pJ}$ for INT8.} The energy consumption formulas for SNNs and ANNs are as follows, 
\begin{equation}
E_{SNNs} = T \times ( fr \times E_{ACs} \times \eta_{ACs} + E_{MACs} \times \eta_{MACs} ),
\end{equation}
\begin{equation}
E_{ANNs} = E_{MACs} \times \eta_{MACs}.
\end{equation}
Here, fr represents the firing rate, T represents the time steps, and \(\eta\) represents the number of operations.

\subsection{Ablation Studies}
\begin{figure}[!t]
\centering
    \includegraphics[width=1.\columnwidth]{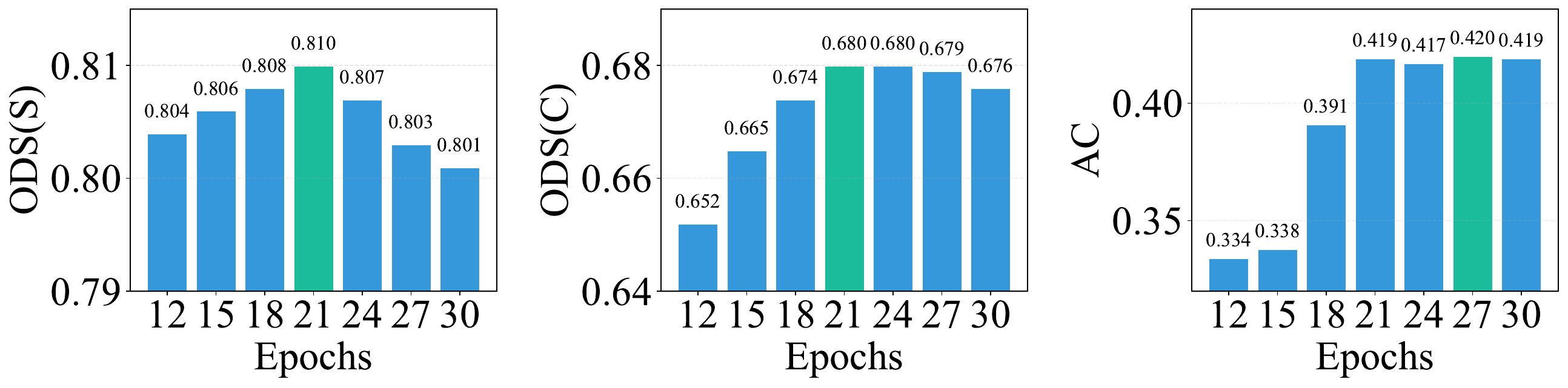}
\caption{\textbf{{Model performance at different epochs during training on the BSDS500 dataset.}}}
\label{fig_hyper}
\end{figure}

In this section, we conduct ablation studies on BSDS500 to evaluate the proposed method. Since AP and OIS show consistent trends with ODS, we only report ODS under both S-Eval and C-Eval.
{As shown in Figure \ref{fig_hyper}, we evaluate prediction performance at different epochs during model training on the BSDS500 dataset, showing that the model converges and achieves optimal performance at epoch 21.}

\begin{table}[tbp]
\centering
\caption{\textbf{The ablation study on MS2ResNet.} 
{The suffixes -N, -W, and -E denote model variants that are narrower following \cite{he_deep_2016}, 1.25$\times$ wider than the baseline, and with two encoder layers replaced by one, respectively.}
The number following the model name indicates the network depth, with the specific configuration of MS2ResNet detailed in Appendix B.}
\resizebox{\columnwidth}{!}{%
\begin{tabular}{lcccccc}
\hline
Backbone & Params & Scale & ODS(S) & ODS(C) & AC & E.(mJ) \\
\hline
VGG19 & 43.33 & 1 & 0.802 & 0.667 & 0.394 & 66.24 \\
MS-ResNet34 & 47.10 & 1 & 0.798 & 0.669 & 0.405 & 86.23 \\
SEW-ResNet34 & 47.10 & 1 & 0.802 & 0.671 & 0.379 & 505.93 \\
EMS-ResNet34 & 46.75 & 1 & 0.799 & 0.670 & 0.395 & 83.75 \\
MDS-ResNet34 & 47.62 & 1 & 0.803 & 0.670 & 0.409 & 79.86 \\
\hline
MS2ResNet42-N & 20.80 & 4 & 0.800 & 0.658 & 0.363 & 33.45 \\
{MS2ResNet42-W} & {68.36} & {4} & {0.808} & {0.682} & {0.413} & {121.33} \\
{MS2ResNet42-E} & {43.58} & {4} & {0.807} & {0.673} & {0.395} & {83.67} \\
\hline
MS2ResNet14 & 28.61 & 4 & 0.795 & 0.660 & 0.381 & 56.61 \\
MS2ResNet26 & 30.63 & 4 & 0.800 & 0.662 & 0.382 & 62.49 \\
\textbf{MS2ResNet42} & \textbf{43.76} & \textbf{4} & \textbf{0.810} & \textbf{0.680} & \textbf{0.419} & \textbf{77.21} \\
MS2ResNet112 & 119.31 & 4 & 0.807 & 0.691 & 0.406 & 108.29 \\
\hline
MS2ResNet42 & 44.90 & 1 & 0.802 & 0.669 & 0.390 & 78.50 \\
MS2ResNet42 & 44.75 & 2 & 0.806 & 0.671 & 0.413 & 77.54 \\
MS2ResNet42 & 42.94 & 8 & 0.809 & 0.679 & 0.416 & 75.85 \\
\hline
\end{tabular}%
}
\label{tab_1}
\end{table}

\subsubsection{The Effectiveness of MS2ResNet}
As shown in rows 1-5 and 11 of Table \ref{tab_1}, we compare MS2ResNet with other spiking backbones. Our backbone significantly outperforms other spiking backbones in terms of ODS(S), ODS(C), and AC metrics, while maintaining slightly lower parameters and energy consumption. 
Notably, for a fair comparison with VGG19 \cite{cordone2022object}, we match the width of other Spiking Residual Networks with it. We find that this width configuration helps enhance model performance. As shown in rows 6 and 11 of Table \ref{tab_1}, when we adjust the network width to follow the original setting in \cite{he_deep_2016}, although the parameters and energy consumption decrease, the performance drops significantly.
{Conversely, when we increase the model width to 1.25$\times$ the baseline (row 7), overfitting occurs, causing both S-Eval and AC to drop while C-Eval remains nearly unchanged.
Additionally, using a single encoding layer instead of two (row 8) decreases all metrics while increasing energy consumption despite parameter reduction, confirming that deeper encoding layers better capture information-rich spike representations.}

In Section \ref{analysis}, we theoretically prove that MS2ResNet avoids gradient vanishing or exploding problems, enabling deep network architectures. Our experimental results in rows 9-12 of Table \ref{tab_1} confirm this, showing significant performance improvements as network depth increases. However, models exceeding 100 layers exhibit overfitting, with increasing C-Eval but decreasing S-Eval and AC. Consequently, we select MS2ResNet42 as our baseline model.
Furthermore, we investigate the optimal number of scales in our multi-scale residual learning strategy. As shown in rows 11 and 13-15, integrating 4 scales achieves the best performance. We can observe that more scales are not always better. Using 8 scales leads to overfitting due to excessive model capacity.

\begin{table}[tbp]
\centering
\setlength{\tabcolsep}{5pt}
\caption{\textbf{The ablation study on quantization error.}}
\begin{tabular}{lcccccc}
\hline
 Neuron & T-D & MDS & ODS(S) & ODS(C) & AC & E.(mJ) \\
\hline
% ReLU & - & $\checkmark$ & 0.802 & 0.667 & 0.368 & 430.07 \\
% \hline
IF & 2-1 & $\checkmark$ & 0.807 & 0.673 & 0.364 & 60.74 \\
LIF & 2-1 & $\checkmark$ & 0.805 & 0.672 & 0.368 & 59.47 \\
PLIF & 2-1 & $\checkmark$ & 0.807 & 0.670 & 0.369 & 61.54 \\
\hline
I-LIF & 1-2 & $\checkmark$ & 0.807 & 0.674 & 0.382 & 58.99 \\
I-LIF & 1-4 & $\times$ & 0.807 & 0.678 & 0.414 & 68.97 \\
\textbf{I-LIF} & \textbf{1-4} & \textbf{$\checkmark$} & \textbf{0.810} & \textbf{0.680} & \textbf{0.419} & \textbf{77.21} \\
I-LIF & 1-8 & $\checkmark$ & 0.809 & 0.680 & 0.414 & 88.06 \\
I-LIF & 2-4 & $\checkmark$ & 0.808 & 0.679 & 0.425 & 136.87 \\
I-LIF & 4-2 & $\checkmark$ & 0.802 & 0.682 & 0.402 & 204.63 \\
\hline
\end{tabular}
\label{tab_2}
\end{table}

\subsubsection{The Influence of Quantization Error}
We experimentally validate our method, which combines I-LIF and MDS, to overcome quantization errors. First, we compare I-LIF with other commonly used neurons including IF, LIF, and PLIF. Since I-LIF converts trained integer values to binary spikes during inference, its $T_{inference} = T \times D$. To leverage the advantages of I-LIF while ensuring fair comparison, we set $T_{inference} = 1 \times 2$ for I-LIF and T = 2 for other neurons. As shown in rows 1-4 of Table \ref{tab_2}, I-LIF effectively mitigates the impact of quantization errors and achieves optimal metric balance with lower energy consumption.
Then we compare the model performance with and without MDS. As shown in rows 5 and 6 of Table \ref{tab_2}, MDS stabilizes the membrane synaptic input distribution, effectively alleviating quantization errors while achieving better performance with lower energy consumption.

As shown in rows 4, 6-9 of Table \ref{tab_2}, we increase T and D in I-LIF. The performance improves with increasing $T_{inference}$ but degrades when $T_{inference} > 4$. This validates our analysis in Section \ref{MS2ResNet}: while alleviating quantization errors reduces edge discontinuity, it weakens the suppression of texture information. At $T_{inference} = 1 \times 4$, the model achieves an optimal balance between boundary line preservation and texture suppression. % As shown in the first row of Table \ref{tab_2}, we also train an ANN version of MS2Edge by replacing all spiking neurons with ReLU. Its performance degrades compared to the SNN version, further validating our analysis that SNNs provide a strong prior for edge detection.

\begin{table}[tbp]
\centering
\setlength{\tabcolsep}{5pt}
\caption{{\textbf{The ablation study on SNN-specific mechanisms.}
The numbers under Weight and Activation denote the bit-width. For SNNs, S represents spike-based activation. For the 4-bit quantization case, since \cite{horowitz20141} does not report $E_{\text{MACs}}$ for 4-bit integer, we use $<$ to indicate that its energy consumption is lower than the 8-bit case.
}}
\begin{tabular}{ccccccc}
\hline
Spike & Weight & Activation & ODS(S) & ODS(C) & AC & E.(mJ) \\
\hline
$\times$ & 32 & 32 & 0.802 & 0.667 & 0.368 & 430.07 \\
$\times$ & 8 & 8 & 0.789 & 0.669 & 0.376 & 68.70 \\
$\times$ & 4 & 4 & 0.687 & 0.501 & 0.102 & $<68.70$ \\
$\times$ & 32 & 8 & 0.790 & 0.671 & 0.377 & 430.07 \\
\textbf{$\checkmark$} & \textbf{32} & \textbf{S} & \textbf{0.810} & \textbf{0.680} & \textbf{0.419} & \textbf{77.21} \\
\hline
\end{tabular}
\label{tab_3}
\end{table}

\subsubsection{The Effectiveness of SNN-Specific Mechanisms}
{
To explicitly quantify the contribution of SNN mechanisms, we conduct comprehensive ablation studies on quantization and spike-driven computation mechanisms.
To evaluate the quantization mechanism separately, we employ quantization methods of ANNs, which can exhibit a similar effect to SNNs.
In implementation, we employ post-training quantization from MQBench \cite{MQBench} with MinMax calibration to quantize the ANN version of MS2Edge, as shown in Table \ref{tab_3}.
After quantizing both activations and weights to 8-bit, S-Eval drops significantly, while C-Eval and AC slightly improve, with reduced energy consumption. 
This demonstrates that quantization effectively enhances edge crispness. 
As for the drop in S-Eval, we believe it is caused by existing quantization techniques (both post-training quantization and quantization-aware training) relying on pre-trained ANN weights \cite{MQBench}, whose performance is fundamentally bounded by the original ANNs, limiting their ability to fully exploit quantization benefits.
% Moreover, unlike SNNs that can be trained end-to-end, quantization methods require pre-training the entire network, incurring substantial computational costs.
When further quantized to 4-bit, all metrics drop dramatically due to excessive quantization errors.
Additionally, since SNNs inherently quantize neural outputs through the spike-driven mechanism, we conduct an activation-only quantization experiment to enable a fair comparison. As shown in row 4 of Table \ref{tab_3}, the results show similar patterns but with no energy reduction, as mixed precision prevents the use of efficient integer arithmetic units.}

{
As SNNs cannot emit floating-point outputs, independent ablation of the spike-driven computation mechanism is impossible.
Therefore, we compare the ANN and SNN versions of MS2Edge to evaluate the overall contribution of both mechanisms. As shown in rows 1 and 5 of Table \ref{tab_3}, when comparing the ANN version with the SNN version, SNNs achieve superior performance across all metrics, with significantly lower energy consumption. These experiments validate our analysis that SNNs provide a strong prior for edge detection in an energy-efficient manner.
} 

\begin{table}[tbp]
\centering
\setlength{\tabcolsep}{5pt}
\caption{\textbf{The effectiveness and different scale of SMSUB.}}
\begin{tabular}{cccccccc}
\hline
SMSUB & Scale & Params & ODS(S) & ODS(C) & AC & E.(mJ) \\
\hline
$\times$ & 0 & 29.79 & 0.807 & 0.675 & 0.383 & 46.11 \\
$\checkmark$ & 1 & 30.34 & 0.806 & 0.674 & 0.374 & 45.35 \\
$\checkmark$ & 2 & 30.93 & 0.806 & 0.669 & 0.378 & 43.51 \\
$\checkmark$ & 3 & 35.21 & 0.807 & 0.674 & 0.382 & 54.60 \\
$\checkmark$ & 4 & 39.48 & 0.807 & 0.675 & 0.411 & 67.54 \\
\textbf{$\checkmark$} & \textbf{5} & \textbf{43.76} & \textbf{0.810} & \textbf{0.680} & \textbf{0.419} & \textbf{77.21} \\
$\checkmark$ & 6 & 48.04 & 0.808 & 0.685 & 0.414 & 82.06 \\
\hline
\end{tabular}
\label{tab_4}
\end{table}

\subsubsection{The Effectiveness of SMSUB}
\begin{figure}[!t]
\centering
    \includegraphics[width=1.\columnwidth]{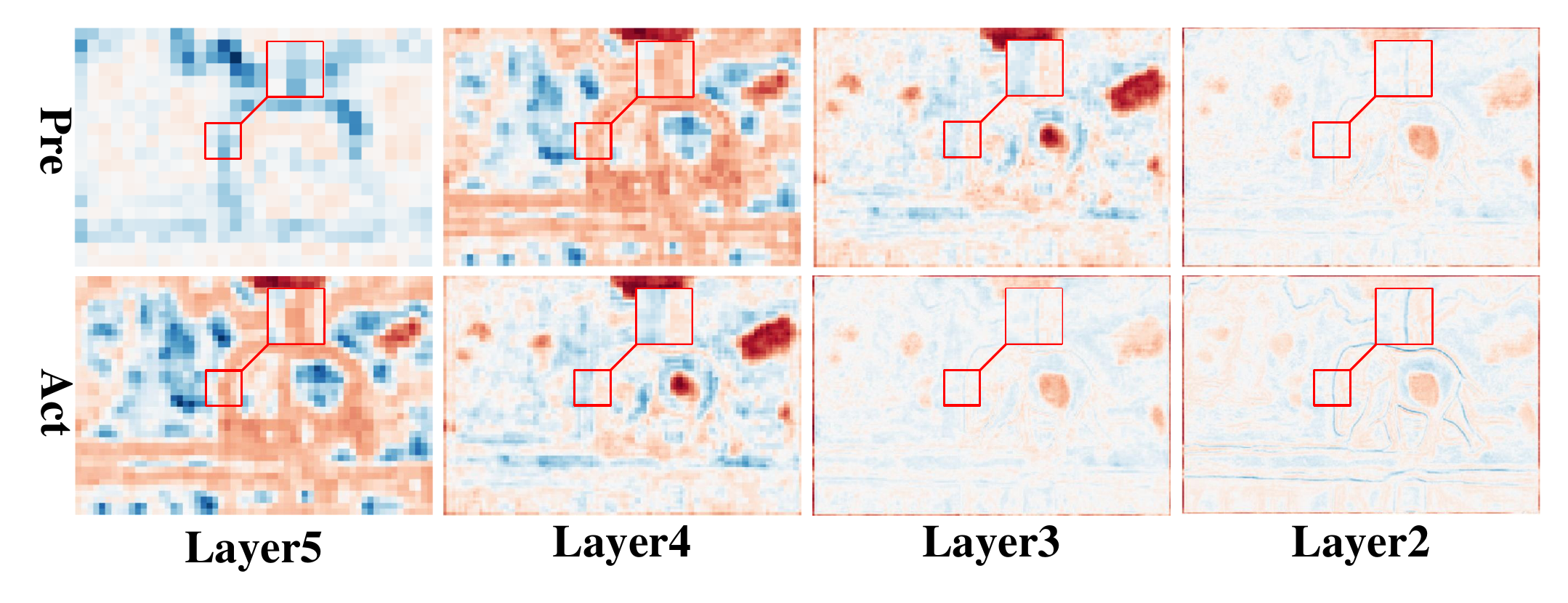}
\caption{\textbf{The feature map of SMSUB.} The top row shows feature maps after nearest-neighbor interpolation but before multi-scale processing, while the bottom row shows feature maps after SMSUB.}
\label{fig_6}
\end{figure}

We verify the effectiveness of SMSUB by replacing it with a 3$\times$3 convolution. As shown in rows 1 and 6 of Table \ref{tab_4}, SMSUB not only ensures comparable energy consumption but also substantially improves edge crispness.
Then, we thoroughly investigate the optimal number of scales in SMSUB. As shown in rows 2-7 of Table \ref{tab_4}, the model achieves the best performance with 5 scales without significantly increasing energy consumption or parameters.  
Additionally, we visualize SMSUB feature maps before and after multi-scale processing in Fig. \ref{fig_6}, which demonstrates that multi-scale processing effectively overcomes the similar response problem.

% Notably, the model performs poorly with only 1 scale, which is equivalent to no multi-scale processing.

\begin{table}[htbp]
\centering
\setlength{\tabcolsep}{5pt}
\caption{\textbf{Comparison of different spiking decoding methods.}}
\begin{tabular}{ccccc}
\hline
Decode Method & ODS(S) & ODS(C) & AC & E.(mJ) \\
\hline
SRD & 0.807 & 0.678 & 0.412 & 82.11 \\
LMPD & 0.809 & 0.675 & 0.406 & 77.53 \\
\textbf{MAD} & \textbf{0.810} & \textbf{0.680} & \textbf{0.419} & \textbf{77.21} \\
\hline
\end{tabular}
\label{tab_5}
\end{table}

\subsubsection{The Effectiveness of MAD}
We compare our proposed MAD with SRD and LMPD. The comparative results are presented in Table \ref{tab_5}. The MAD method not only maintains comparable energy consumption but also achieves higher ODS(S), ODS(C), and AC scores compared to these alternatives.

\begin{table}[htbp]
\centering
\setlength{\tabcolsep}{5pt}
\caption{\textbf{Comparison of different loss functions.}}
\begin{tabular}{ccccc}
\hline
Loss & ODS(S) & ODS(C) & AC & E.(mJ) \\
\hline
BCE \cite{zhang_generalized_2018} & 0.808 & 0.671 & 0.348 & 74.55 \\
WCE \cite{Xie_2015_ICCV} & 0.810 & 0.676 & 0.265 & 77.99 \\
FL \cite{Lin_2017_ICCV} & 0.807 & 0.675 & 0.344 & 74.34\\
Mix \cite{liu2024generating} & 0.808 & 0.683 & 0.393 & 75.84 \\
\textbf{HFL} \cite{liu2025cycle} & \textbf{0.810} & \textbf{0.680} & \textbf{0.419} & \textbf{77.21} \\
\hline
\end{tabular}
\label{tab_6}
\end{table}

\subsubsection{Comparison of Different Loss Functions}
Since this is the pioneering work of SNNs in edge detection, we evaluate different loss functions, as shown in Table \ref{tab_6}.
The Hybrid Focal Loss Function (HFL) improves performance with lower energy consumption. We attribute this to its effective constraint from both image-level and pixel-level information, which overcomes the imbalanced sample distribution and enhances classification accuracy between edge and non-edge pixels.

\subsection{Comparison with SOTA Methods}
\label{Comparison Experiments}

\begin{table*}[htbp]
	\caption{\textbf{Quantitative comparison on BSDS500 dataset. $\dagger$ indicates training with extra PASCAL VOC Context data. $\ddagger$ indicates the multi-scale testing. $^\&$ means generating multi-granularity edges. $\star$ denotes the cited GPU speed, and $*$ denotes the inference speed tested on A40 GPU. The best and second-best performances are marked in \textcolor{red}{red} and \textcolor{blue}{blue}, respectively.}}
	\label{BSDS500_Tab}
	\centering
    \begin{tabular}{c|c|c|ccc|ccc|c|c|c|c}
        \hline
        & \multirow{2}{*}{Methods}& \multirow{2}{*}{Pretrained}& \multicolumn{3}{c|}{S-Eval}& \multicolumn{3}{c|}{C-Eval}& \multirow{2}{*}{AC}& \multirow{2}{*}{Params(M)}& \multirow{2}{*}{FPS}& \multirow{2}{*}{E.(mJ)}\\
        \cline{4-9}
        &&&ODS& OIS& AP& ODS& OIS& AP& & &\\
        \hline
        \multirow{5}{*}{\rotatebox{90}{Traditional}}
        &Canny \cite{canny2009computational}& No& 0.611& 0.676& 0.520& -& -& -& -& -& -& -\\
        &gPb-UCM \cite{arbelaez2010contour}& No& 0.729& 0.755& 0.745& -& -& -& -& -& -& -\\
        &CED-ADM \cite{li2021color}& No& 0.731& 0.760& 0.605& -& -& -& -& -& -& -\\
        &SE \cite{dollar_fast_2014}& No& 0.743& 0.764& 0.800& -& -& -& -& -& -& -\\
        &MES \cite{sironi2015projection}& No& 0.756& 0.776& 0.756& -& -& -& -& -& -& -\\
        \hline
        \multirow{13}{*}{\rotatebox{90}{ANN-based}}
        &HED \cite{Xie_2015_ICCV}& Yes& 0.788& 0.808& 0.840& 0.576& 0.591& -& 0.215& 14.70& 30$\star$& 151.17\\
        &RCF \cite{Liu_2017_CVPR}& Yes& 0.798& 0.815& -& 0.585& 0.604& -& 0.189& 14.80& 30$\star$& 137.89\\
        &CED \cite{wangDeepCrispBoundaries2017}& Yes& 0.794& 0.811& -& 0.642& 0.656& -& 0.207& 14.90& -& -\\
        &LPCB \cite{Deng_2018_ECCV}& Yes& 0.800& 0.816& -& 0.693& 0.700& -& -& 17.70& 30$\star$& 147.50\\
        &BDCN \cite{he_bidirectional_2019}& Yes& 0.806& 0.826& 0.847& 0.636& 0.650& -& 0.233& 16.30& 44$\star$& 385.57\\
        &FCL-Net \cite{xuan2022fcl}& Yes& 0.807& 0.822& -& 0.652& 0.666& 0.570& -& 16.46& 8$*$& -\\
        &EDTER \cite{pu_edter_2022}& Yes& 0.824& 0.841& 0.880& \textcolor{blue}{0.698}& \textcolor{blue}{0.706}& -& 0.288& 468.84& 2.1$\star$& 763.60\\
        &MuGE$^\&$ \cite{zhou2024muge}& Yes& \textcolor{red}{0.850}& \textcolor{red}{0.856}& \textcolor{red}{0.896}& 0.671& 0.678& 0.713& 0.294& 93.9& 28$*$& 167.64\\
        &RankED \cite{cetinkaya2024ranked}& Yes& 0.824& 0.840& \textcolor{blue}{0.895}& 0.611& 0.619& 0.653& 0.290& 114.4& -& 400.16\\
        \cline{2-13}
        &DexiNed \cite{soria2023dense}& No& 0.729& 0.745& 0.583& -& -& -& -& 35.2& -& 177.67\\
        &PiDiNet \cite{Su_2021_ICCV}& No& 0.789& 0.803& -& 0.578& 0.587& -& 0.202& \textcolor{red}{0.71}& \textcolor{red}{69$*$}& \textcolor{blue}{78.87}\\
        & CHRNet \cite{elharrouss_refined_2023}& No & 0.787& 0.788& 0.801& -& -& -& -& -& -& -\\
        &PEdger$\dagger\ddagger$ \cite{fu2023practical}& No& 0.821& 0.841& -& -& -& -& 0.333& \textcolor{blue}{0.73}& \textcolor{blue}{66$*$}& 174.26\\
        &CPD-Net \cite{liu2025cycle}& No& 0.792& 0.811& 0.798& 0.660& 0.669& 0.707& 0.339& 9.75& 48$*$& 352.52\\
        \hline
        \multirow{3}{*}{\rotatebox{90}{Ours}}
        &MS2Edge& No& 0.810& 0.832& 0.835& 0.680& 0.690& 0.728& \textcolor{red}{0.419}& \multirow{3}{*}{43.76}& \multirow{3}{*}{47$*$}& \textcolor{red}{77.21}\\
        &MS2Edge$\dagger$& No& 0.823& 0.842& 0.843& \textcolor{red}{0.700}& \textcolor{red}{0.713}& \textcolor{red}{0.766}& \textcolor{blue}{0.388}& & & 80.15\\
        &MS2Edge$\dagger\ddagger$& No& \textcolor{blue}{0.827}& \textcolor{blue}{0.846}& 0.854& 0.669& 0.679& 0.730& 0.299& & & 80.15\\
        \hline
    \end{tabular}
\end{table*}

\subsubsection{BSDS500} 
\begin{figure*}[htbp]
    \centering
    \subfloat[BSDS500\label{fig:PR_a}]{
        \includegraphics[width=0.45\textwidth]{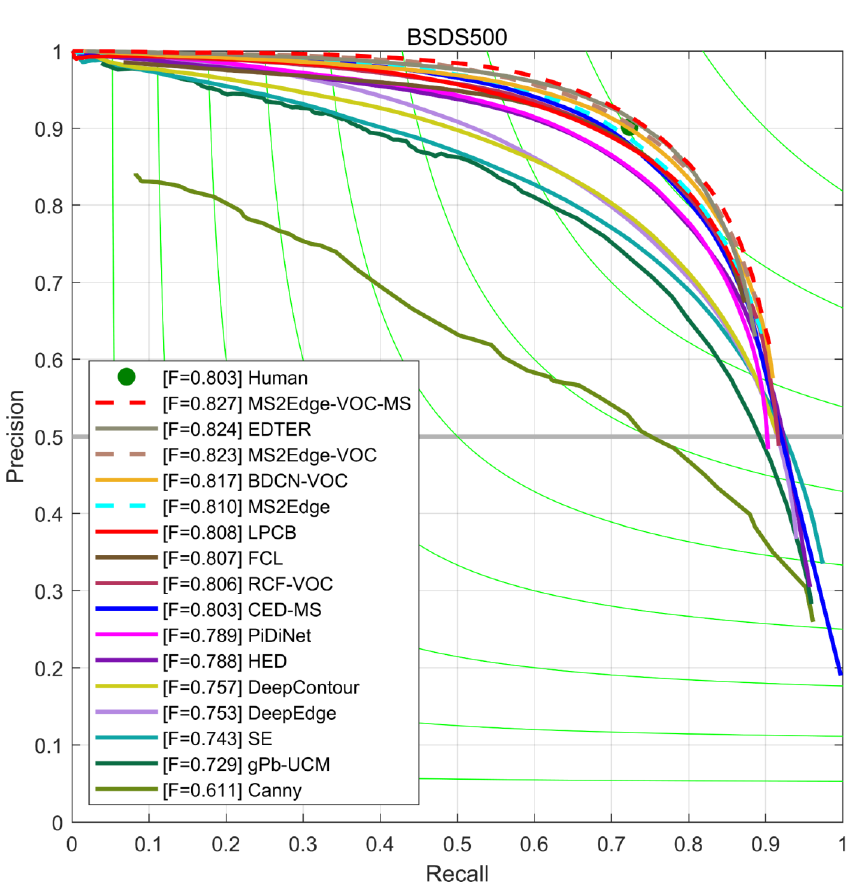}
    }
    \hfill
    \subfloat[NYUDv2\label{fig:PR_b}]{
        \includegraphics[width=0.45\textwidth]{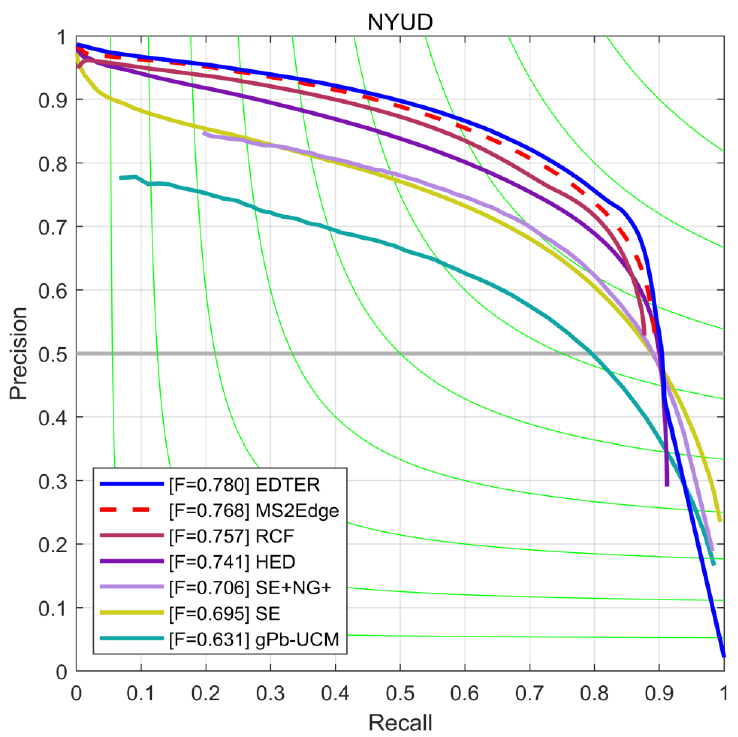}
    }
    \caption{Precision-Recall curves on BSDS500 and NYUD-V2 dataset, respectively.}
    \label{PR_curves}
\end{figure*}

\begin{figure}[htbp]
    \centering
    \includegraphics[width=1.\columnwidth]{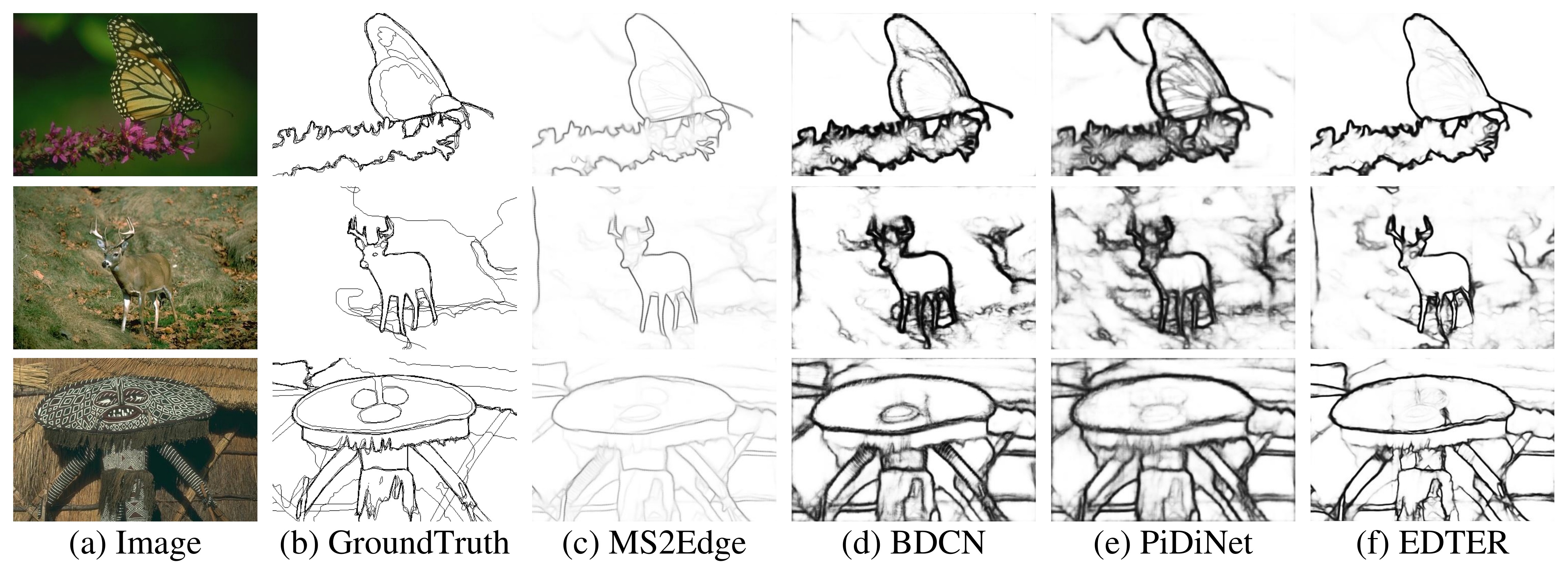}
    \caption{Some examples from SOTA methods on BSDS500.}
    \label{BSDS500}
\end{figure}

It can be seen from Table \ref{BSDS500_Tab} and Fig. \ref{PR_curves} that our MS2Edge achieves high performance while maintaining ultralow energy consumption among SOTA methods. 
It is worth noting that our FPS measurements for SNNs are conducted on GPUs. Unlike ANNs, SNNs have not received comparable optimization for GPU inference.
For S-Eval, though slightly inferior to MuGE, our model achieves competitive performance (ODS=0.827, OIS=0.846, AP=0.854) with only 80.15 mJ, making it substantially more efficient than MuGE's 167.64 mJ. Notably, MuGE's superior performance stems directly from its multi-granularity strategy whereas our method employs a single-granularity framework.
For the C-Eval metrics, our MS2Edge significantly outperforms recent SOTA methods including EDTER, MuGE, and RankED, achieving impressive performance (ODS=0.700, OIS=0.713, AP=0.766). 
Notice that many existing methods rely heavily on the pre-training backbone for prior knowledge, while our MS2Edge achieves superior performance without any pre-training requirements, making its training procedure more straightforward and resource-efficient. 
Compared with other models without pre-trained backbones such as PiDiNet and PEdger, we demonstrate significant advantages in terms of accuracy while still maintaining advantages in energy consumption. This result underscores the strong prior that SNN characteristics contribute to edge detection.
The visualization of our results from Fig. \ref{BSDS500} also shows cleaner edge predictions with fewer artifacts in non-edge regions.

\begin{table*}[htbp]
	\caption{\textbf{Quantitative comparison results on NYUDv2 dataset. The best and second-best performances are marked in \textcolor{red}{red} and \textcolor{blue}{blue}, respectively.}}
	\label{NYUD_Tab}
	\centering
	\begin{tabular}{c|ccc|ccc|ccc|c|c}
	\hline
	  \multirow{3}{*}{Methods}& \multicolumn{9}{c|}{S-Eval}& \multirow{3}{*}{AC}& \multirow{3}{*}{E.(mJ)}\\
        \cline{2-10}
        & \multicolumn{3}{c|}{RGB}& \multicolumn{3}{c|}{HHA}& \multicolumn{3}{c|}{RGB-HHA}& &\\
        \cline{2-10}
        & ODS& OIS& AP& ODS& OIS& AP& ODS& OIS& AP& &\\
	  \hline
	  gPb-UCM \cite{arbelaez2010contour}& 0.631& 0.661& 0.562& -& -& -& -& -& -& -& -\\
	  SE \cite{dollar_fast_2014}& 0.695& 0.708& 0.719& -& -& -& -& -& -& -& -\\
        SE+NG+ \cite{gupta2014learning}& 0.706& 0.734& 0.549& -& -& -& -& -& -& -& -\\
	  \hline
	  HED \cite{Xie_2015_ICCV}& 0.720& 0.734& 0.734& 0.682& 0.695& 0.702& 0.746& 0.761& 0.786& -& 219.70\\
        RCF \cite{Liu_2017_CVPR}& 0.729& 0.742& -& 0.705& 0.715& -& 0.757& 0.771& -& -& 216.43\\
        LPCB \cite{Deng_2018_ECCV}& 0.739& 0.754& -& 0.707& 0.719& -& 0.762& 0.778& -& -& 232.31\\
        BDCN \cite{he_bidirectional_2019}& 0.748& 0.763& 0.770& 0.707& 0.719& \textcolor{red}{0.731}& 0.765& 0.781& \textcolor{blue}{0.813}& 0.162& 603.54\\
        EDTER \cite{pu_edter_2022}& \textcolor{blue}{0.774}& \textcolor{blue}{0.789}& \textcolor{blue}{0.797}& 0.703& 0.718& \textcolor{blue}{0.727}& \textcolor{red}{0.780}& \textcolor{red}{0.797}& \textcolor{red}{0.814}& 0.195& 763.60\\
        RankED \cite{cetinkaya2024ranked}& \textcolor{red}{0.780}& \textcolor{red}{0.793}& \textcolor{red}{0.826}& -& -& -& -& -& -& -& 400.16\\
        \hline
        DexiNed \cite{soria2023dense}& 0.658& 0.674& 0.556& -& -& -& -& -& -& -& 277.84\\
        PiDiNet \cite{Su_2021_ICCV}& 0.733& 0.747& -& 0.715& 0.728& -& 0.756& 0.773& -& 0.173& \textcolor{blue}{121.81}\\
        CHRNet \cite{elharrouss_refined_2023} & 0.729& 0.745& -& \textcolor{red}{0.718}& \textcolor{red}{0.731}& -& 0.750& 0.774& -& - & - \\
        PEdger \cite{fu2023practical}& 0.742& 0.757& -& -& -& -& -& -& -& -& 1079.38\\
        CPD-Net \cite{liu2025cycle}& 0.735& 0.750& 0.690& \textcolor{blue}{0.717}& \textcolor{blue}{0.730}& 0.679& 0.760& 0.775& 0.770& \textcolor{blue}{0.223}& 555.22\\
        \hline
        MS2Edge& 0.752& 0.767& 0.734& 0.702& 0.718& 0.660& \textcolor{blue}{0.770}& \textcolor{blue}{0.784}& 0.777& \textcolor{red}{0.239}& \textcolor{red}{76.22} \\

        \hline
        \noalign{\vspace{8pt}}
        \hline

        % C-Eval 部分
        \multirow{3}{*}{Methods}& \multicolumn{9}{c|}{C-Eval}& \multirow{3}{*}{AC}& \multirow{3}{*}{E.(mJ)}\\
        \cline{2-10}
        & \multicolumn{3}{c|}{RGB}& \multicolumn{3}{c|}{HHA}& \multicolumn{3}{c|}{RGB-HHA}& &\\
        \cline{2-10}
        & ODS& OIS& AP& ODS& OIS& AP& ODS& OIS& AP& &\\
	  \hline
	  HED \cite{Xie_2015_ICCV}& 0.387& 0.404& -& 0.335& 0.350& -& 0.368& 0.384& -& -& 219.70\\
        RCF \cite{Liu_2017_CVPR}& 0.395& 0.412& -& 0.333& 0.348& -& 0.374& 0.397& -& -& 216.43\\
        BDCN \cite{he_bidirectional_2019}& 0.414& 0.439& -& 0.347& 0.367& -& 0.375& 0.392& -& 0.162& 603.54\\
        EDTER \cite{pu_edter_2022}& 0.416& 0.438& 0.372& \textcolor{red}{0.394}& \textcolor{red}{0.410}& \textcolor{red}{0.348}& \textcolor{blue}{0.402}& \textcolor{blue}{0.420}& \textcolor{blue}{0.368}& 0.195& 763.60\\
        RankED \cite{cetinkaya2024ranked}& \textcolor{red}{0.477}& \textcolor{red}{0.490}& \textcolor{red}{0.474}& -& -& -& -& -& -& -& 400.16\\
        \hline
        PiDiNet \cite{Su_2021_ICCV}& 0.399& 0.424& 0.344& -& -& -& -& -& -& 0.173& \textcolor{blue}{121.81}\\
        CPD-Net \cite{liu2025cycle}& 0.453& 0.468& 0.415& 0.358& 0.371& 0.289& 0.398& 0.417& 0.362& \textcolor{blue}{0.223}& 555.22\\
        \hline
        MS2Edge& \textcolor{blue}{0.472}& \textcolor{blue}{0.485}& \textcolor{blue}{0.431}& \textcolor{blue}{0.374}& \textcolor{blue}{0.386}& \textcolor{blue}{0.303}& \textcolor{red}{0.421}& \textcolor{red}{0.438}& \textcolor{red}{0.383}& \textcolor{red}{0.239}& \textcolor{red}{76.22} \\
        \hline
	\end{tabular}
\end{table*}

\subsubsection{NYUDv2}
\begin{figure}[htbp]
    \centering
    \includegraphics[width=1.\columnwidth]{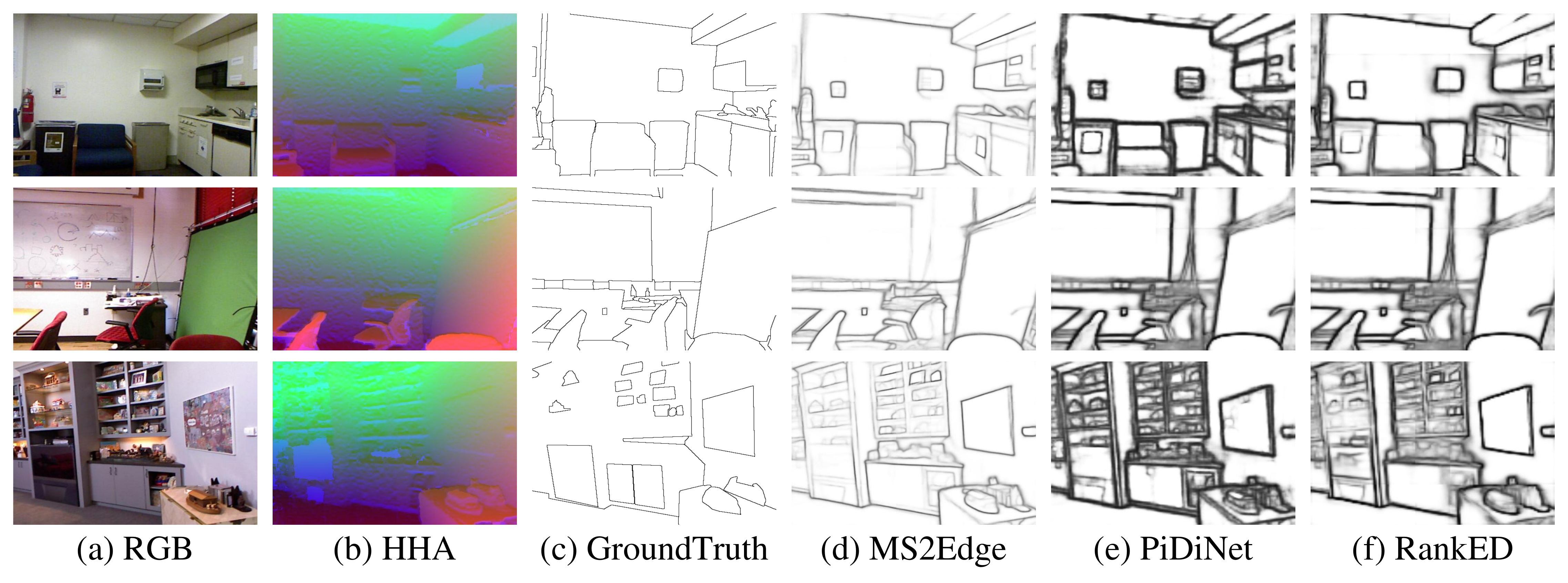}
    \caption{Some examples from SOTA methods on NYUDv2.}
    \label{NYUDv2}
\end{figure}

The comparison results on the NYUDv2 dataset are shown in Table \ref{NYUD_Tab}. 
For S-Eval, on RGB images, our MS2Edge significantly outperforms other methods without pre-trained backbones, achieving scores of ODS=0.752, OIS=0.767, and AP=0.734, while guaranteeing $\frac{1}{2}$ or even lower energy consumption.
On RGB-HHA fusion, MS2Edge demonstrates strong cross-modality learning capability, achieving ODS=0.770, OIS =0.784, and AP=0.777. It not only achieves SOTA performance among methods without pre-training but also ranks second compared to methods with pre-training while using only $\frac{1}{10}$ of the parameters and the energy consumption compared to EDTER.
Additionally, in C-Eval metrics, MS2Edge exhibits robust performance on RGB images with ODS=0.472, OIS=0.485, and AP=0.431. It achieves the best performance on RGB-HHA fusion for all methods with ODS=0.421, OIS=0.438, and AP=0.383. 
In terms of AC score, our method demonstrates an even more substantial lead, surpassing the second-place competitor by a margin of $\frac{1}{3}$.
The visual results from Fig. \ref{NYUDv2} exhibit consistent characteristics with those on BSDS500. 

\subsubsection{BIPED}
As shown in Table \ref{BIPED_Tab}, our MS2Edge demonstrates superior performance across all evaluation metrics on the BIPED dataset. The multi-scale testing variant, MS2Edge$\ddagger$, achieves SOTA performance with ODS=0.902, OIS=0.908, and AP=0.929, surpassing all top methods including the recent EDTER (ODS=0.893, OIS=0.898) and Li et al. (ODS=0.890, OIS=0.894, AP=0.902). Notably, our model achieves this exceptional performance without any pre-training requirements. The base MS2Edge model also demonstrates impressive performance with ODS=0.901, OIS=0.907, and AP=0.922, and achieves a remarkable AC score of 0.499, substantially higher than other SOTA methods including EDTER (AC=0.260) and CPD-Net (AC=0.426). 

\begin{table}[htbp]
    \caption{\textbf{Quantitative comparison results on BIPED dataset. $\ddagger$ refers to multi-scale testing. The best and second-best performances are marked in \textcolor{red}{red} and \textcolor{blue}{blue}, respectively.}}
    \label{BIPED_Tab}
    \centering
    \begin{tabular}{c|ccc|c}
        \hline
	  Methods& ODS& OIS& AP& AC\\
	  \hline
        SED \cite{akbarinia2018feedback}& 0.717& 0.731& 0.756& -\\
        HED \cite{Xie_2015_ICCV}& 0.829& 0.847& 0.869\\
	  RCF \cite{Liu_2017_CVPR}& 0.849& 0.861& 0.906& -\\
        BDCN \cite{he_bidirectional_2019}& 0.890& 0.899& \textcolor{red}{0.934}&-\\
        CED-ADM \cite{li2021color}& 0.810& 0.835& 0.869& -\\
        % CATS \cite{huan2021unmixing}& 0.887& 0.892& 0.817& -\\
        FCL-Net \cite{xuan2022fcl}& 0.895& 0.900& -& -\\
        % TEED \cite{soria2023tiny}& 0.828& 0.842& -\\
        EDTER \cite{pu_edter_2022}& 0.893& 0.898& -& 0.260\\
        % DiffusionEdge \cite{ye2024diffusionedge}& \textcolor{blue}{0.899}& 0.901& -& \textcolor{red}{0.849}\\
        % Li et al. \cite{li2024pixel}& 0.890& 0.894& 0.902& 0.318\\
        \hline
        DexiNed \cite{soria2023dense}& 0.895& 0.900& 0.927& 0.295\\
        PiDiNet \cite{Su_2021_ICCV}& 0.868& 0.876& -& 0.232\\
        CPD-Net \cite{liu2025cycle}& 0.896& 0.901& 0.929& \textcolor{blue}{0.426}\\
	  \hline
        MS2Edge& \textcolor{blue}{0.901}& \textcolor{blue}{0.907}& 0.922& \textcolor{red}{0.499}\\
        MS2Edge$\ddagger$& \textcolor{red}{0.902}& \textcolor{red}{0.908}& \textcolor{blue}{0.929}& 0.348\\
        % LUS-Net& 0.902& 0.908& 0.912& \multirow{2}{*}{70.41M}\\
        % LUS-Net$\ddagger$& \textbf{0.903}& \textbf{0.909}& 0.930\\
	\hline
    \end{tabular}
\end{table}

\begin{table}[htbp]
    \caption{\textbf{Quantitative comparison results on PLDU and PLDM dataset. The best and second-best performances are marked in \textcolor{red}{red} and \textcolor{blue}{blue}, respectively.}}
    \label{PLDUAV_Tab}
    \centering
    \begin{tabular}{c|cc|cc}
        \hline
	  \multirow{2}{*}{Methods}& \multicolumn{2}{c|}{PLDU dataset}& \multicolumn{2}{c}{PLDM dataset}\\
        \cline{2-5}
        & ODS& OIS& ODS& OIS\\
	  \hline
        Canny \cite{canny2009computational}& 0.466& 0.643& 0.796& 0.866\\
%         PMI \cite{isola2014crisp}& 0.535& 0.622& 0.641& 0.752\\
        SE \cite{dollar_fast_2014}& 0.850& 0.898& 0.351& 0.340\\
        % LSD \cite{gromponevongioiLSDFastLine2010}& 0.593& 0.593& 0.796& 0.796\\
        % Gestalt Grouping \cite{ipol.2018.194}& 0.629& 0.629& 0.808& 0.808\\
        HED \cite{Xie_2015_ICCV}& 0.905& 0.927& 0.859& 0.883\\
	  RCF \cite{Liu_2017_CVPR}& 0.907& 0.931& 0.865& 0.893\\
        Zhang et al. \cite{zhang2019detecting}& 0.914& 0.938& \textcolor{blue}{0.888}& \textcolor{blue}{0.902}\\
        \hline
        PiDiNet \cite{Su_2021_ICCV}& 0.964& 0.971& 0.809& 0.818\\
        % PEdger \cite{fu2023practical}\\
        CPD-Net \cite{liu2025cycle}& \textcolor{blue}{0.966}& \textcolor{blue}{0.974}& 0.828& 0.844\\
	  \hline
        MS2Edge& \textcolor{red}{0.979}& \textcolor{red}{0.986}& \textcolor{red}{0.949}& \textcolor{red}{0.969}\\
	\hline
    \end{tabular}
\end{table}

\subsubsection{PLDU and PLDM}
Table \ref{PLDUAV_Tab} results on PLDU and PLDM datasets show MS2Edge's exceptional performance in power line detection, demonstrating its strong generalization ability.
On the PLDU dataset, our method achieves outstanding scores with ODS=0.979 and OIS=0.986, significantly outperforming previous SOTA methods including Zhang et al. (ODS=0.914, OIS=0.938) and CPD-Net (ODS=0.966, OIS=0.974). 
Similarly, on the PLDM dataset, MS2Edge achieves superior performance with ODS=0.949 and OIS=0.969, showing substantial improvements over the closest competitor Zhang et al. (ODS=0.888, OIS=0.902). 
MS2Edge outperforms previous best methods by 6-7\% in ODS scores across both PLDU and PLDM datasets, demonstrating robust performance in diverse power line detection scenarios.

\section{Conclusion}\label{conclusion}
In this paper, we propose MS2Edge, the first SNN-based edge detection method, to address two major challenges in edge detection.
First, we construct an SNN-based network, demonstrating that its inherent mechanisms can naturally provide strong edge detection priors in an energy-efficient manner.
Second, we propose a novel SNN backbone, MS2ResNet. Through multi-scale residual learning strategy, it effectively resolves the sparse edge discontinuity problem caused by SNN quantization errors and significantly enhances edge localization capability.
Furthermore, we propose SMSUB to handle convolution similar response issues for better detail reconstruction during upsampling, and introduce MAD for more effective edge map decoding.
Experimental results demonstrate that MS2Edge achieves SOTA performance on BSDS500, NVUDV2, and BIPED edge detection datasets, as well as on PLDU and PLDM datasets for power line detection, while achieving significantly lower energy consumption compared to existing ANN-based approaches.
This work provides a new perspective for future edge detection methods while also inspiring the application of SNNs to more edge detection downstream tasks, offering high-performance and energy-efficient solutions.

However, we overly rely on the spike-driven characteristics of SNNs without conducting a lightweight design, thus resulting in limited energy advantages over the most efficient ANN-based methods. Therefore, in future work, we plan to explore more lightweight SNN architectures to expand the energy-efficiency gap and extend MS2Edge to more edge detection downstream tasks, providing high-performance and energy-efficient methods across diverse domains.

\bibliographystyle{IEEEtran}
\bibliography{references}

\appendices
\section{Proof of Gradient Norm Equality}\label{proof}
\subsection{Proof of Proposition 1}

\noindent\textbf{Proposition 1.}
\textit{For MS2Block1 and MS2Block2, if \(\beta_{decay} = 0\) and the block output at each time step $t$ follows \(\mathbf{x^t} \sim \mathcal{N}(0, 1)\), then each block satisfies: \(\phi(\mathbf{J_{j}^{t}(J_{j}^{t})^{T}}) = \frac{1}{\alpha_{2}^{t,j-1}}\).}

\noindent\rule{\columnwidth}{0.5pt}

\noindent\textbf{Theorem 2.}
\textbf{(Theorem 4.2. in \cite{chen2020comprehensive})} 
\textit{Given \(\mathbf{J}:=\sum_j \mathbf{J_j}\), where \({\mathbf{J_j}}\) is a series of independent random matrices. If at most one matrix in \({\mathbf{J_j}}\) is not a central matrix, we have}
\begin{equation}
\phi \left( \mathbf{JJ}^T \right) = \sum_j \phi \left( \mathbf{J_j}\mathbf{J_j}^T \right).
\end{equation}

\noindent\textbf{Lemma 1.}
\textbf{(Lemma A.11. in \cite{chen2020comprehensive})}
\textit{Let \(\mathbf{f(x)}\) be a transform whose Jacobian matrix is $\mathbf{J}$. If $\mathbf{f(x) = Jx}$, then we have}
\begin{equation}
E\left[\frac{||\mathbf{f(x)}||_2^2}{len(\mathbf{f(x)})}\right] = \phi\left(\mathbf{J}\mathbf{J}^T\right)E\left[\frac{||\mathbf{x}||_2^2}{len(\mathbf{x})}\right].
\end{equation}

\noindent\textbf{Proposition 3.}
\textbf{{(Proposition 5.5. in \cite{chen2020comprehensive})}}
\textit{For a serial neural network \(\mathbf{f(x)}\) composed of general linear transforms and its input-output Jacobian matrix is
\(\mathbf{J}\), we have}
\begin{equation}
E\left[\frac{||\mathbf{f(x)}||_2^2}{len(\mathbf{f(x)})}\right] = \phi\left(\mathbf{J}\mathbf{J}^T\right)E\left[\frac{||\mathbf{x}||_2^2}{len(\mathbf{x})}\right].
\end{equation}

When \(\beta_{decay} = 0\), as shown in Eqs. 10 and 11 of the main paper, gradient computation is independent for each time step, allowing us to analyze the network at each $t$ independently. We define \(\mathbf{J_j^t}\) as the Jacobian matrix of the j-th block at time step $t$. For MS2Block1 and MS2Block2, each consists of residual and shortcut paths. We define the Jacobian matrices of these two paths at time step $t$ as \(\mathbf{J_{res}^t}\) and \(\mathbf{J_{sc}^t}\), respectively. Both blocks share the same residual path but differ in their shortcut paths. 

First, we analyze the residual path, which includes convolution, tdBN, and I-LIF operations. While convolution and tdBN have been proven to satisfy Definition 2 in previous studies \cite{chen2020comprehensive, zheng2021going}, we focus our analysis on I-LIF. When \(\beta_{decay} = 0\), the neuron model degenerates as
\begin{equation}\label{ILIF2_beta0}
\mathbf{s}^{t,j} = \mathit{Clip} \left( \mathit{round}(\mathbf{x}^{t,j}), 0, D \right).
\end{equation}

During training, I-LIF uses rectangular windows as the surrogate function, thus its Jacobian matrix is as follows.
\begin{equation}
J_{j,pq}^t = 
\begin{cases}
1, & \text{if } p = q \text{ and } 0 \leq x_{j,pq}^t \leq D \\
0, & \text{if } p = q \text{ and } (x_{j,pq}^t < 0 \text{ or } x_{j,pq}^t > D) \\
0, & \text{if } p \neq q,
\end{cases}
\end{equation}
where $p$ and $q$ denote the row and column coordinates. During forward propagation, this surrogate function is equivalent to
\begin{equation}
\mathbf{f(x^t)} = 
\begin{cases}
\mathbf{x^t},  & \text{if } 0 \leq \mathbf{x^t} \leq D \\
0,  & \text{otherwise}.
\end{cases}
\end{equation}
For this surrogate function, it is evident that $\mathbf{f(x^t) = Jx^t}$. By Lemma 1 and Definition 2, this surrogate function satisfies the definition of general linear transform. Since the original function employs the gradient of this surrogate function during backpropagation, I-LIF can be regarded as a general linear transform.

From the above derivation, the residual path is composed of serial general linear transforms. Therefore, by Proposition 3, denoting $\alpha_2$ as the second-order moment, we can derive:
\begin{equation}
\alpha_2^{t,j,res} = \phi(\mathbf{J_{res}^t}(\mathbf{J_{res}^t})^T)\alpha_2^{t,j-1},
\end{equation}
\begin{equation}\label{J_res}
\phi(\mathbf{J_{res}^t}(\mathbf{J_{res}^t})^T) = \frac {\alpha_2^{t,j,res}} {\alpha_2^{t,j-1}}.
\end{equation}

For the shortcut path, MS2Block1 and MS2Block2 differ in that MS2Block2 replaces the 1$\times$1 convolution with a 3$\times$3 convolution for better spatial information integration during downsampling. Both convolutions satisfy the general linear transform. The other transforms, tdBN and I-LIF, also satisfy the general linear transform as discussed above. Therefore, by Proposition 3, we can derive:
\begin{equation}
\alpha_2^{t,j,sc} = \phi(\mathbf{J_{sc}^t}(\mathbf{J_{sc}^t})^T)\alpha_2^{t,j-1},
\end{equation}
\begin{equation}\label{J_sc}
\phi(\mathbf{J_{sc}^t}(\mathbf{J_{sc}^t})^T) = \frac {\alpha_2^{t,j,sc}} {\alpha_2^{t,j-1}}.
\end{equation}

According to Eqs. \ref{J_res}, \ref{J_sc}, and Theorem 2, for MS2Block1 and MS2Block2, we have:

\begin{equation}
\begin{split}
\phi(\mathbf{J_{MSBlock1}^t}(\mathbf{J_{MSBlock1}^t}&)^T) = \phi(\mathbf{J_{MSBlock2}^t}(\mathbf{J_{MSBlock2}^t})^T) \\
&= \phi(\mathbf{J_{sc}^t}(\mathbf{J_{sc}^t})^T) + \phi(\mathbf{J_{res}^t}(\mathbf{J_{res}^t})^T) \\
&= \frac {\alpha_2^{t,j,sc} + \alpha_2^{t,j,res}} {\alpha_2^{t,j-1}}.
\end{split}
\end{equation}

Therefore, when the block output at each time step $t$ follows \(\mathbf{x^t} \sim \mathcal{N}(0, 1)\), MS2Block1 and MS2Block2 satisfy \(\phi(\mathbf{J_{j}^{t}(J_{j}^{t})^{T}}) = \frac{1}{\alpha_{2}^{t,j-1}}\).

\subsection{Proof of Proposition 2}
\noindent\textbf{Proposition 2.}
\textit{For MS2ResNet, if all conditions stated in Proposition 1 hold and the $2^{nd}$ moment of the encoding layer output can be controlled at each time step $t$, then \(\phi(\mathbf{J^t(J^t)^T}) \approx 1\). Consequently, the training of MS2ResNet can avoid gradient vanishing or exploding problems.}

\noindent\rule{\columnwidth}{0.5pt}

In MS2ResNet, we incorporate MS-Block when the model deepens to reduce model complexity and facilitate gradient propagation. For MS-Block, according to the derivation in \cite{su2023deep}, we have
\begin{equation}
\phi(\mathbf{J_{MS-Block}^t}(\mathbf{J_{MS-Block}^t})^T) = \frac{\alpha_{2}^{t,j-1} + 1}{\alpha_{2}^{t,j-1}}.
\end{equation}
If each MS2Block1 and MS2Block2 output follows \(\mathbf{x^t} \sim \mathcal{N}(0, 1)\), therefore for the MS-Block following the MS2Block, then we can derive
\begin{equation}\label{j_ms}
\phi(\mathbf{J_{MS-Block}^t}(\mathbf{J_{MS-Block}^t})^T) = 2.
\end{equation}
If it is a cascade of MS2Blocks, we can also infer the current $\phi(\mathbf{J^t_j}(\mathbf{J^t_j})^T)$ based on the preceding $\alpha_{2}^{t,j-1}$. Subsequently, we can conduct independent derivations based on Theorem 1, Proposition 1, and Eq. \ref{j_ms} for MS2ResNet with different configurations. For example, for MS2ResNet42, we can derive
\begin{equation}
\phi(\mathbf{J^t}(\mathbf{J^t})^T) = \frac{144}{\alpha_{2}^{t,0}},
\end{equation}
where $\alpha_{2}^{t,0}$ is the second-order moment of encoding layer output. 
Therefore, if $\alpha_{2}^{t,0}$ can be controlled, then \(\phi(\mathbf{J^t(J^t)^T}) \approx 1\).

Therefore, according to Definition 1, MS2ResNet achieves Block Dynamical Isometry at each time step t. Consequently, $E \left[ \left|\left|\Delta\theta_i^t\right|\right|^2 \right]$ maintains a stable value, neither approaching zero nor infinity. Moreover, since $\Delta\theta_i = \sum_{t=1}^T \Delta\theta_i^t$, we can derive $E \left[ \left|\left|\Delta\theta_i\right|\right|^2 \right]$ as follows.
\begin{equation}\label{theata1}
\begin{split}
E \left[ \left|\left|\Delta\theta_i\right|\right|^2 \right] &= E \left[ (\Delta\theta_i)^2 \right] \\
&= E \left[ (\sum_{t=1}^T \Delta\theta_i^t)^2 \right] \\
&= E \left[ (\sum_{t=1}^T \sum_{s=1}^T (\Delta\theta_i^t \cdot \Delta\theta_i^s)) \right]
\end{split}
\end{equation}

Assuming that $\Delta\theta_i^t$ is \textit{i.i.d} for each time step t, when $i = s$, we have
\begin{equation}
E \left[ (\Delta\theta_i^t \cdot \Delta\theta_i^s) \right] = E \left[ (\Delta\theta_i^t)^2 \right].
\end{equation}
When $i \neq s$, we have
\begin{equation}
\begin{split}
E \left[ (\Delta\theta_i^t \cdot \Delta\theta_i^s) \right] &= E \left[ (\Delta\theta_i^t) \right] \cdot E \left[ (\Delta\theta_i^s) \right] \\
&= (E \left[ (\Delta\theta_i^t) \right])^2.
\end{split}
\end{equation}
Substituting the above derivation into Eq. \ref{theata1}, we have
\begin{equation}
E \left[ (\Delta\theta_i^t \cdot \Delta\theta_i^s) \right] = T \cdot E \left[ (\Delta\theta_i^t)^2 \right] + ( T^2 - T ) \cdot (E \left[ (\Delta\theta_i^t) \right])^2.
\end{equation}
Since the learning rate of the network is typically less than 1e-3, we have $(E \left[ (\Delta\theta_i^t) \right])^2 \approx 0$. Therefore, we can derive
\begin{equation}
\begin{split}
E \left[ \left|\left|\Delta\theta_i\right|\right|^2 \right ] &\approx T \cdot E \left[ (\Delta\theta_i^t)^2 \right] \\
&\approx T \cdot E \left[ \left|\left|\Delta\theta_i^t\right|\right|^2 \right ].
\end{split}
\end{equation}
As discussed above, since $E \left[ \left|\left|\Delta\theta_i^t\right|\right|^2 \right ]$ is stable, $E \left[ \left|\left|\Delta\theta_i\right|\right|^2 \right ]$ is also stable. Therefore, the training of MS2ResNet can avoid gradient vanishing or exploding problems.

\begin{table*}
\centering
\caption{\textbf{The Structure of MS2ResNet for BSDS500.} The Output Size column indicates the spatial dimensions and number of channels in the output feature map at each stage.}
\begin{adjustbox}{width=0.8\textwidth,center}
\begin{tabular}{c|c|c|c|c|c}
\hline
\rule{0pt}{2ex} Stage & Output Size & MS2ResNet14 & MS2ResNet26 & MS2ResNet42 & MS2ResNet112 \rule[-1ex]{0pt}{0pt} \\
\hline
\rule{0pt}{4ex} Encoding & 480$\times$320, 64 & \multicolumn{4}{c}{Encoding Block $\times$ 2} \rule[-2ex]{0pt}{0pt} \\
\hline
\rule{0pt}{5ex} Conv1 & 240$\times$160, 128 &  
MS2Block2 $\times$ 1  & 
\makecell{MS2Block2 $\times$ 1 \\ MS2Block1 $\times$ 1} &
\makecell{MS2Block2 $\times$ 1 \\ MS2Block1 $\times$ 1 \\MS-Block $\times$ 1 } &
\makecell{MS2Block2 $\times$ 1 \\ MS-Block $\times$ 1 \\MS2Block1 $\times$ 1 }
\rule[-3ex]{0pt}{0pt} \\
\hline
\rule{0pt}{5ex} Conv2 & 120$\times$80, 256 &
MS2Block2 $\times$ 1  & 
\makecell{MS2Block2 $\times$ 1 \\ MS2Block1 $\times$ 1} &
\makecell{MS2Block2 $\times$ 1 \\ MS-Block $\times$ 1 \\MS2Block1 $\times$ 1 \\ MS-Block $\times$ 1} &
\makecell{MS2Block2 $\times$ 1 \\ MS-Block $\times$ 3 \\MS2Block1 $\times$ 1 \\ MS-Block $\times$ 3}
\rule[-3ex]{0pt}{0pt} \\
\hline
\rule{0pt}{5ex} Conv3 & 60$\times$40, 512 &
MS2Block2 $\times$ 1  & 
\makecell{MS2Block2 $\times$ 1 \\ MS2Block1 $\times$ 1} &
\makecell{MS2Block2 $\times$ 1 \\ MS-Block $\times$ 2 \\MS2Block1 $\times$ 1 \\ MS-Block $\times$ 2} &
\makecell{MS2Block2 $\times$ 1 \\ MS-Block $\times$ 15 \\MS2Block1 $\times$ 1 \\ MS-Block $\times$ 15}
\rule[-3ex]{0pt}{0pt} \\
\hline
\rule{0pt}{5ex} Conv4 & 30$\times$20, 512 &
MS2Block2 $\times$ 1  & 
\makecell{MS2Block2 $\times$ 1 \\ MS2Block1 $\times$ 1} &
\makecell{MS2Block2 $\times$ 1 \\ MS2Block1 $\times$ 1 \\MS-Block $\times$ 1} &
\makecell{MS2Block2 $\times$ 1 \\ MS-Block $\times$ 3 \\MS2Block1 $\times$ 1 \\ MS-Block $\times$ 3}
\rule[-3ex]{0pt}{0pt} \\
\hline
\end{tabular}
\end{adjustbox}
\label{tab_Structure}
\end{table*}

\section{More Implementation Details}\label{more}
The architecture of MS2ResNet is shown in Table \ref{tab_Structure}. As the model goes deeper, we also partially utilize the MS-Block \cite{su2023deep} to facilitate gradient propagation while preserving the functionality of the MS2Block.

The process of multi-scale testing consists of three main steps: First, Image Pyramid Construction: We create an image pyramid comprising three resolutions of the input image (0.5×, 1.0×, and 2.0×) using bilinear interpolation. Second, Multi-scale Processing: Each scaled image is independently processed through MS2Edge. The resulting edge maps are then restored to the original input resolution. Third, Edge Map Fusion: The three restored edge maps are averaged to produce a final fused edge map.
% Additionally, to overcome the impact of multiple annotators in edge detection datasets, we adopt the label-smoothing method proposed in \cite{fu2023practical} to synthesize label information. 

\begin{figure*}[htb]
    \centering
    \includegraphics[width=\textwidth]{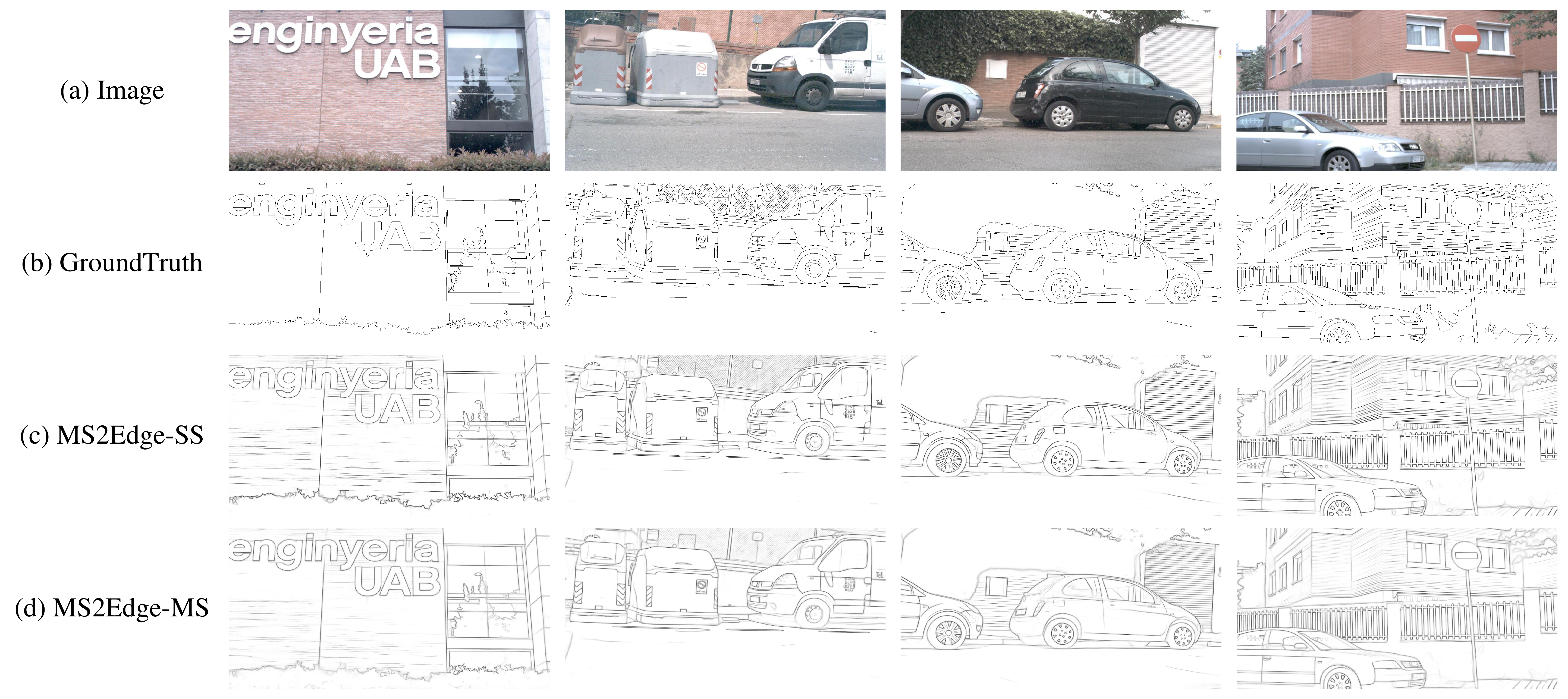}
    \caption{Some examples from MS2Edge on BIPED. MS2Edge-SS indicates the predictions with single-scale testing, and MS2Edge-MS indicates the predictions with multi-scale testing.}
    \label{fig_biped}
\end{figure*}

\begin{figure*}[hb]
    \centering
    \includegraphics[width=1.\textwidth]{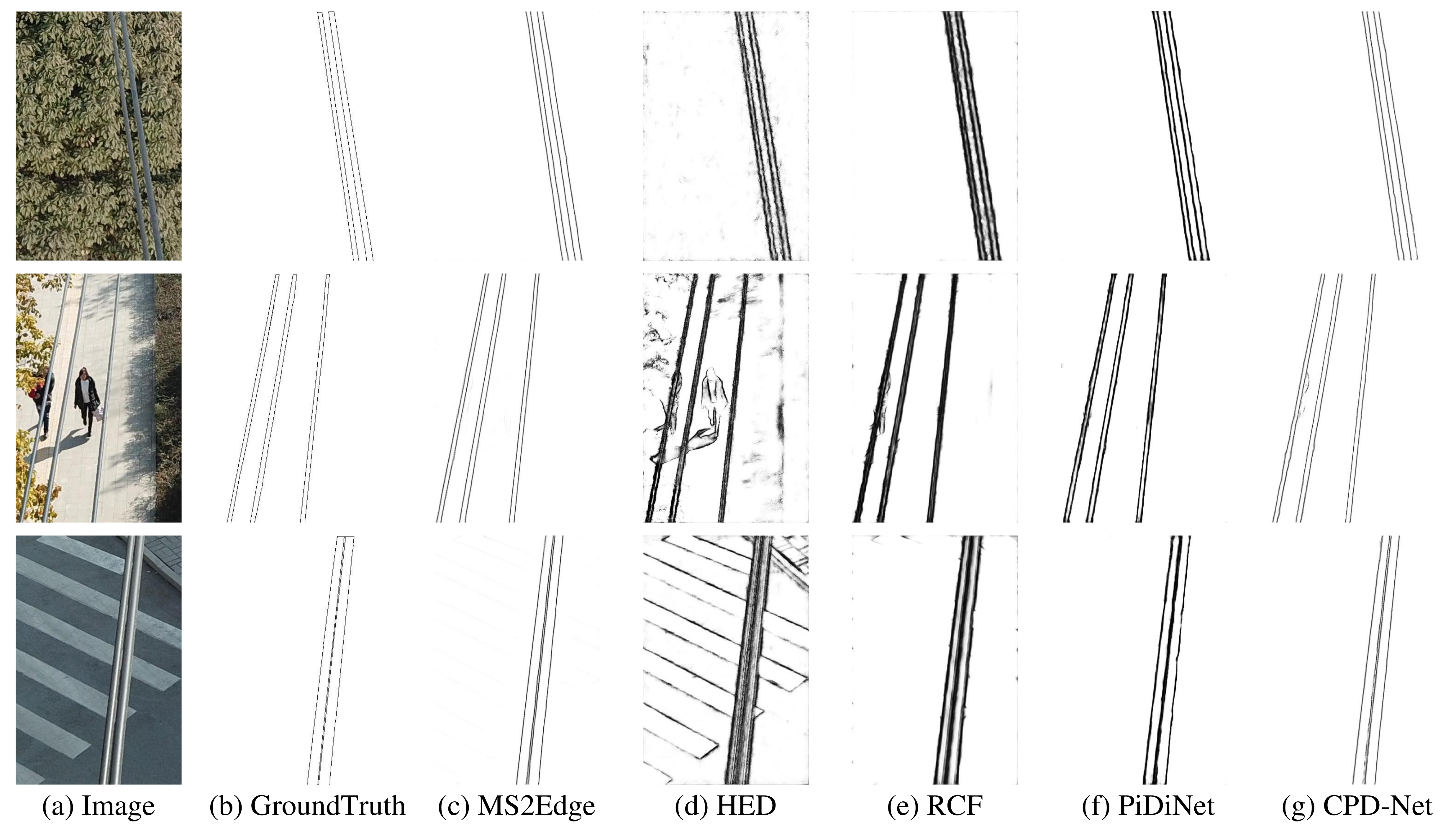}
    \caption{Some examples from different SOTA methods on PLDU dataset.}
    \label{fig_PLDU}
\end{figure*}

\begin{figure*}[hb]
    \centering
    \includegraphics[width=1.\textwidth]{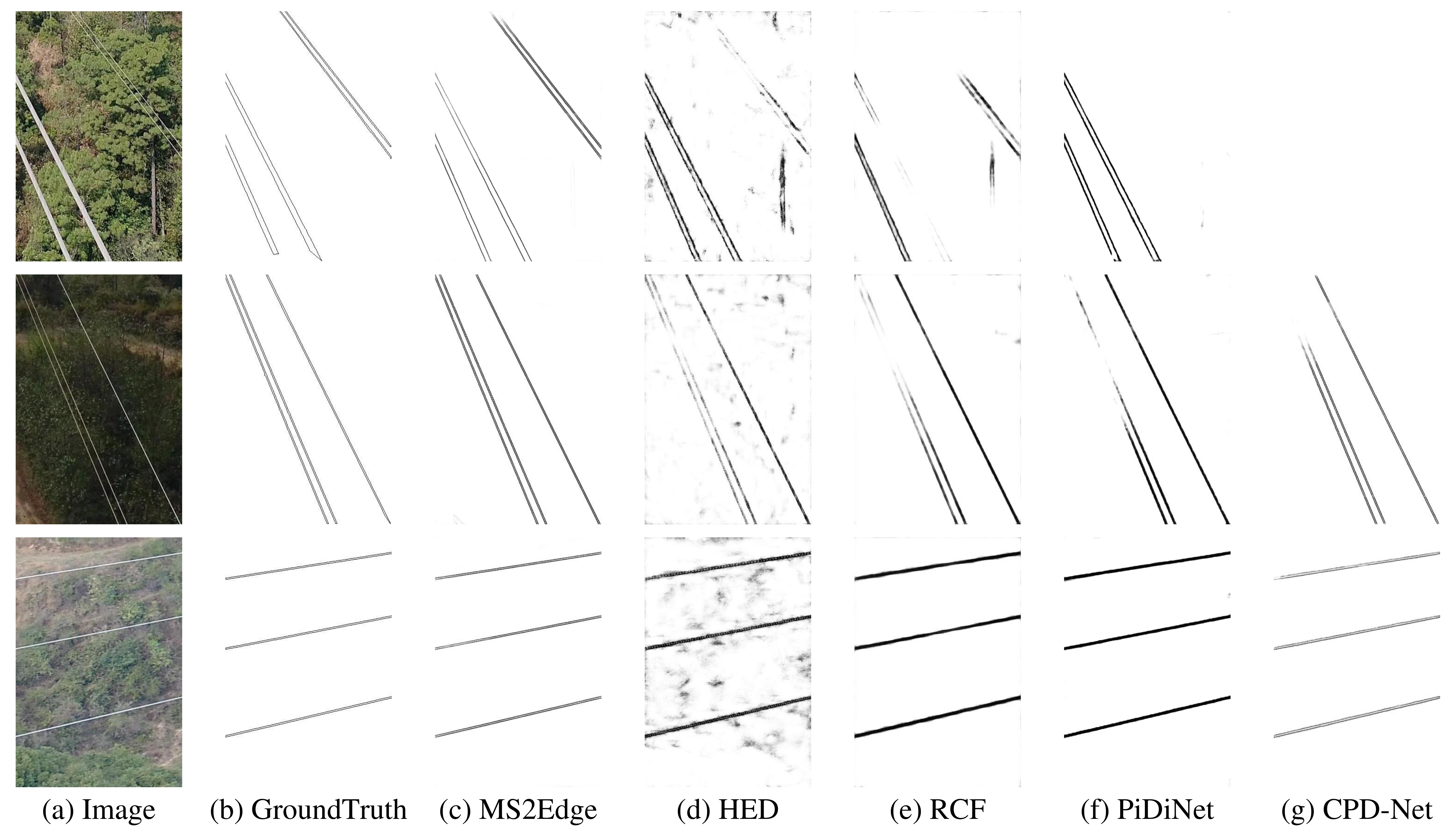}
    \caption{Some examples from different SOTA methods on PLDM dataset.}
    \label{fig_PLDM}
\end{figure*}

\section{More Visualization}
\label{more_vis}
We show more visualized results on BIPED \cite{soria2023dense}, PLDU, and PLDM \cite{zhang2019detecting}, which are shown in Fig. \ref{fig_biped}, Fig. \ref{fig_PLDU}, and Fig. \ref{fig_PLDM}, respectively. These visualized results demonstrate that the MS2Edge can generate clean and crisp edge maps, which are consistent with those of BSDS500 \cite{arbelaez2010contour} and NYUDv2 \cite{silberman_indoor_2012}.

\end{document}